\documentclass[manuscript,screen]{acmart}

\AtBeginDocument{%
  }

\setcopyright{acmlicensed}
\copyrightyear{2018}
\acmYear{2018}
\acmDOI{XXXXXXX.XXXXXXX}
\acmConference[]{ACM Computing Surveys}{June 03--05,
  2018}{Woodstock, NY}
\acmISBN{978-1-4503-XXXX-X/2018/06}
\usepackage{graphicx}
\usepackage[edges]{forest}   
\usepackage{xcolor, colortbl}

\usepackage{amssymb} %
\usepackage{amsmath}
\usepackage{hyperref}
\usepackage{natbib}
\usetikzlibrary{arrows.meta,calc,fit}
\usepackage{booktabs}
\usepackage{multirow}
\usepackage{makecell}    
\usepackage{adjustbox}
\usepackage{array}
\usepackage{tabularx}   
\usepackage{ragged2e}   
\newcolumntype{L}[1]{>{\RaggedRight\arraybackslash}p{#1}} 
\newcolumntype{Y}{>{\RaggedRight\arraybackslash}X}        
\begin{document}

\title[RL Meets LLMs: A Survey of Advancements and Applications Across the LLM Lifecycle]{Reinforcement Learning Meets Large Language Models: A Survey of Advancements and Applications Across the LLM Lifecycle}


\author{Keliang Liu}
\authornote{Both authors contributed equally to this research.}
\orcid{0009-0002-2728-5109}   
\email{klliu25@m.fudan.edu.cn}
\affiliation{%
  \institution{Fudan University}
  \city{Shanghai}
  \country{China}
}

\author{Dingkang Yang}
\authornotemark[1] 
\authornote{Project leader.}
\orcid{0000-0003-1829-5671}  
\email{dkyang20@fudan.edu.cn}
\affiliation{%
  \institution{Fudan University}
  \city{Shanghai}
  \country{China}
}
\affiliation{%
  \institution{ByteDance SAIL Team}
  \city{Shanghai}
  \country{China}
}

\author{Ziyun Qian}
\orcid{0009-0007-9800-1253} 
\email{zyqian22@m.fudan.edu.cn}
\affiliation{%
  \institution{Fudan University}
  \city{Shanghai}
  \country{China}
}

\author{Weijie Yin}
\orcid{0009-0007-9891-6621} 
\email{yinwj2021@163.com}
\affiliation{%
  \institution{ByteDance SAIL Team}
  \city{Shanghai}
  \country{China}
}

\author{Yuchi Wang}
\orcid{0009-0000-1906-8820} 
\email{wangyuchi@link.cuhk.edu.hk}
\author{Hongsheng Li}
\orcid{0000-0002-2664-7975}
\email{hsli@ee.cuhk.edu.hk}
\affiliation{%
  \institution{The Chinese University of Hong Kong, MMLab}
  \city{Hongkong}
  \country{China}
}


\author{Jun Liu}
\orcid{0000-0002-4365-4165} 
\email{j.liu81@lancaster.ac.uk}
\affiliation{
  \institution{Lancaster University}
  \city{Lancaster}
  \country{UK}
}

\author{Peng Zhai}
\authornotemark[3]  
\orcid{0000-0002-1374-7969} 
\email{pzhai@fudan.edu.cn}
\affiliation{%
  \institution{Fudan University}
  \city{Shanghai}
  \country{China}
}

\author{Yang Liu}
\authornotemark[3]  
\orcid{0000-0002-1312-0146}  
\email{yangliu@cs.toronto.edu}
\affiliation{%
  \institution{Tongji University}
  \city{Shanghai}
  \country{China}
}
\affiliation{%
  \institution{The University of Toronto}
  \city{Toronto}
  \country{Canada}
}

\author{Lihua Zhang}
\authornote{Corresponding authors.}
\orcid{0000-0003-0467-4347}  
\email{lihuazhang@fudan.edu.cn}
\affiliation{%
  \institution{Fudan University}
  \city{Shanghai}
  \country{China}
}



\renewcommand{\shortauthors}{Keliang Liu \textit{et al.}}

\begin{abstract}
In recent years, training methods centered on Reinforcement Learning (RL) have markedly enhanced the reasoning and alignment performance of Large Language Models (LLMs), particularly in understanding human intents, following user instructions, and bolstering inferential strength. Although existing surveys offer overviews of RL augmented LLMs, their scope is often limited, failing to provide a comprehensive summary of how RL operates across the full lifecycle of LLMs. We systematically review the theoretical and practical advancements whereby RL empowers LLMs, especially Reinforcement Learning with Verifiable Rewards (RLVR). First, we briefly introduce the basic theory of RL. Second, we thoroughly detail application strategies for RL across various phases of the LLM lifecycle, including pre-training, alignment fine-tuning, and reinforced reasoning. In particular, we emphasize that RL methods in the \textit{``reinforced reasoning''} phase serve as a pivotal driving force for advancing model reasoning to its limits. Next, we collate existing datasets and evaluation benchmarks currently used for RL fine-tuning, spanning human-annotated datasets, AI-assisted preference data, and program-verification-style corpora. Subsequently, we review the mainstream open-source tools and training frameworks available, providing clear practical references for subsequent research. Finally, we analyse the future challenges and trends in the field of RL-enhanced LLMs. This survey aims to present researchers and practitioners with the latest developments and frontier trends at the intersection of RL and LLMs, with the goal of fostering the evolution of LLMs that are more intelligent, generalizable, and secure.
\end{abstract}

\ccsdesc[500]{General and reference~Surveys and overviews}
\ccsdesc[500]{Computing methodologies~Natural language processing}
\ccsdesc[500]{Computing methodologies~Reinforcement learning}

\keywords{Reinforcement Learning, Large Language Models, Reasoning, Alignment, Reinforcement Learning from Human Feedback.}


\maketitle

\section{Introduction}~\label{sec1}
Large Language Models such as ChatGPT~\cite{ChatGPT} have risen rapidly demonstrating remarkable performance across various tasks, including general dialogue~\cite{bang2023multitask}, code generation~\cite{liu2023your}, and mathematical reasoning~\cite{frieder2023mathematical}, and have gradually become essential cornerstones for interactive artificial intelligence systems~\cite{liang2025llm,lei2025large,chen2024efficiency,chen2024detecting,yang2024pediatricsgpt,yang2025improving}. Despite their broad generalization capabilities, current LLMs still struggle with crucial shortcomings: they often fail to reliably capture nuanced human intentions and can produce misleading or unsafe outputs~\cite{bender2021dangers, bommasani2021opportunities, kenton2021alignment, weidinger2021ethical, tamkin2021understanding, gehman2020realtoxicityprompts}. Moreover, several recent studies~\cite{mirzadeh2024gsm, huang2023large, shojaee2025illusion} have indicated that the reasoning capabilities of LLMs still exhibit substantial shortcomings. Therefore, effectively aligning the generative capabilities of LLMs with human preferences, values, and specific task requirements, as well as enhancing their reasoning abilities for addressing complex problems, has emerged as one of the significant challenges in current LLM research. In response, RL has been introduced as a powerful framework to address these challenges by directly optimizing model behavior through interactive feedback and reward signals. Table \ref{tab:rl_comparison} shows the performance improvement of typical models after being trained with RL compared to their baselines. 
\begin{table}[htbp]
\centering
\caption{This table compares representative models trained with RL against their baseline counterparts, showing that RL substantially enhances the performance of foundation models and underscoring the critical importance of reinforcement learning. Among them, Magistral Small-SC\(^{*}\) and Magistral Small-RL\(^{\#}\) refer to Magistral Small-24B-Starting Checkpoint and the result of this model trained only through reinforcement learning, respectively.}
\label{tab:rl_comparison}
\renewcommand{\arraystretch}{1.4}
\resizebox{\textwidth}{!}{%
\begin{tabular}{lcccccc}
\toprule
\textbf{Model / Benchmark} & \textbf{AIME2024} & \textbf{GPQA-Diamond} & \textbf{LiveCodeBench} & \textbf{MATH-500} & \textbf{MMLU} & \textbf{SWE-benchVerified} \\
\midrule
DeepSeek-V3~\cite{liu2024deepseek} & 39.2 & 59.1 & 36.2 & 90.2 & 88.5 & 42.0 \\
DeepSeek-R1-Zero~\cite{guo2025deepseek} & 71.0 \textbf{(+31.8)} & 73.3 \textbf{(+14.2)} & 50.0 \textbf{(+13.8)} & 95.9 \textbf{(+5.7)} & -- & -- \\
DeepSeek-R1~\cite{guo2025deepseek} & 79.8 \textbf{(+40.6)} & 71.5 \textbf{(+12.4)} & 65.9 \textbf{(+29.7)} & 97.3 \textbf{(+7.1)} & 90.8 \textbf{(+2.3)} & 49.2 \textbf{(+7.2)} \\
Magistral Small-SC$^{*}$~\cite{rastogi2025magistral} & 32.2 & 63.4 (GPQA, +SFT) & 22.7 (v5) & 93.2 (+SFT) & -- & -- \\
Magistral Small-RL$^{\#}$~\cite{rastogi2025magistral} & 65.8 \textbf{(+33.6)} & 68.8 (GPQA, \textbf{+5.4}) & 46.4 (v5, \textbf{+23.7}) & 95.4 \textbf{(+2.2)} & -- & -- \\
GPT-4o-0513~\cite{hurst2024gpt} & 9.3 & 49.9 & 32.9 & 74.6 & 87.2 & 38.8 \\
OpenAI-o1-1217~\cite{jaech2024openai} & 79.2 \textbf{(+70.2)} & 75.7 \textbf{(+25.8)} & 63.4 \textbf{(+30.5)} & 96.4 \textbf{(+21.8)} & 91.8 \textbf{(+4.6)} & 48.9 \textbf{(+10.1)} \\
\bottomrule
\end{tabular}%
}
\end{table}

Since the seminal introduction of Reinforcement Learning from Human Feedback (RLHF) by Ouyang \textit{et al.}~\cite{ouyang2022training}, RL-based fine-tuning has become a cornerstone method for improving LLM alignment with human instructions and preferences. By leveraging human evaluative feedback or learned reward models, RLHF enables models to iteratively adjust their outputs toward more preferred and helpful responses, going beyond what supervised training alone can achieve. Building on the success of RLHF for alignment, researchers have more recently begun to apply RL paradigms to bolster reasoning capabilities. Notably, starting around 2024, a series of advanced LLMs demonstrated substantial improvements on complex reasoning tasks (\textit{e.g.}, in mathematics and programming) by employing test-time or post-training RL techniques. High-profile examples include OpenAI’s o1 system~\cite{jaech2024openai}, Anthropic’s Claude 3.7/4~\cite{anthropic2025claude}, DeepSeek R1~\cite{guo2025deepseek}, the Kimi K1.5~\cite{team2025kimi}, and Qwen 3~\cite{yang2025qwen3} \textit{etc.}, all of which integrate reinforcement-driven reasoning strategies during inference. These successes suggest that reinforcement learning, when applied at the inference or post-training stage, can unlock new problem-solving abilities in LLMs beyond their pre-trained knowledge. A key innovation underlying these recent advances is the paradigm of Reinforcement Learning with Verifiable Rewards (RLVR)~\cite{lambert2024tulu, guo2025deepseek, yang2025qwen3}, which augments the standard RL loop with objective, automatically verifiable reward signals, such as programmatic checks or proofs of correctness on the model’s output. By rewarding an LLM for producing outputs that pass rigorous correctness tests (\textit{e.g.}, unit tests for code or theorem verifications for math), RLVR directly incentivises the model to generate reliably correct and logically sound solutions. This approach has been a driving force behind the aforementioned reasoning improvements, effectively pushing models to reason through multi-step problems until a verifiable correct result is found. 
Nevertheless, the integration of RL into LLM training and usage raises several open questions and limitations. First, it remains under debate to what extent RLVR truly expands the LLM’s reasoning capabilities beyond what was learned during pre-training~\cite{yue2025does, wu2025invisible, zhao2025echo}. Second, there is no clear consensus on how different RL techniques should be best applied at various stages of the LLM lifecycle, ranging from pre-training and instruction alignment to post-training inference optimization. Third, practical issues of data curation and optimization strategy in RL remain challenging: \textit{e.g.}, constructing high-quality reward datasets via human preference labels, AI-assistant preferences, or programmatic rewards and choosing appropriate RL algorithms such as policy gradients versus reward model optimization are non-trivial design decisions. Finally, the question of how to implement RL fine-tuning efficiently at scale without destabilizing the model’s performance is still not fully resolved.

\definecolor{pre}{HTML}{F8CECC}     
\definecolor{align}{HTML}{FCE4D6}  
\definecolor{rlvr}{HTML}{DAE8FC}    
\definecolor{db}{HTML}{E8F5E9}      
\definecolor{tool}{HTML}{FFF9C4}    
\definecolor{rootcolor}{HTML}{F9F7ED} 

\tikzstyle{my-box}=[
    rectangle,
    rounded corners,
    text opacity=1,
    minimum height=1.5em,
    minimum width=6em,
    inner sep=3pt,
    align=center,
    fill opacity=.3,
    thick,
]
\tikzstyle{leaf}=[
    my-box,
    fill=gray!3,
    text=black,
    align=left,
    text width=59em,
    font=\huge,
    inner xsep=3pt,
    inner ysep=4pt,
]

\begin{figure*}[!t]
    \centering
    \resizebox{\textwidth}{!}{%
        \begin{forest}
            forked edges,
            for tree={
                grow=east,
                reversed=true,
                anchor=base west,
                parent anchor=east,
                child anchor=west,
                base=center,
                font=\huge,
                rectangle,
                draw=gray,
                rounded corners,
                align=left,
                text centered,
                minimum width=5em,
                edge+={gray!60, thick},
                s sep=4pt,
                inner xsep=3pt,
                inner ysep=3pt,
                thick,
                fit=band,
            },
            where level=0{
                folder,
                grow'=0,
                fill=rootcolor,
                font=\Huge,
                child anchor=west,
                parent anchor=south west,
                anchor=west,
                calign=first,
                yshift=10pt,
            }{},
            where level=1{ text width=21em }{},
            where level=2{ text width=24em }{},
            where level=3{ text width=26em }{},
            [ {\quad\quad\textbf{A Taxonomy of RL Enhanced LLMs}\qquad\qquad}
                [\textbf{Pre‑training (\S\ref{sec3.1})}, fill=pre
                    [\textbf{RL in Pre‑training Loop (\S\ref{sec3.1})}, fill=pre
                        [Reinforcement Pre‑Training~\citep{dong2025reinforcement}{,}
                         OctoThinker~\citep{wang2025octothinker}{,}
                         Visual Pre‑Training~\citep{ghosh2025visual}{,}
                         { \textit{etc.}}
                         , leaf]
                    ]
                ]
                [\textbf{Alignment (\S\ref{sec3})}, fill=align
                    [\textbf{Classic Algorithms (\S\ref{sec3.2})}, fill=align
                        [\textbf{RLHF and Reward Modeling}, fill=align
                            [InstructGPT~\citep{ouyang2022training}{,}
                             Bai \textit{\textit{et al.}}~\citep{bai2022training}{,}
                             Text2Reward~\citep{xie2023text2reward}{,}
                             Xiong \textit{et al.}~\citep{xiong2023iterative}{,}
                             Eureka~\citep{ma2023eureka}{,}\\
                             Kwon \textit{et al.}~\citep{kwon2023reward}{,}
                             Wang \textit{et al.}~\cite{wang2025beyond}{,}
                             Fu \textit{et al.}~\cite{fu2025reward}{,}
                             Miao \textit{et al.}~\cite{miao2024inform, miao2025energy}{,}
                             { \textit{etc.}}
                             , leaf]
                        ]
                        [\textbf{Preference Optimization}, fill=align
                            [DPO~\citep{rafailov2023direct}{,}
                             KTO~\citep{ethayarajh2024kto}{,}
                             $\Psi$PO~\citep{azar2024general}{,}
                             Smaug~\citep{pal2024smaug}{,}
                             $\beta$-DPO~\citep{NEURIPS2024_ea888178}{,}
                             SPO~\citep{swamy2024minimaximalist}{,}
                             ORPO~\citep{hong2024orpo}{,}
                             { \textit{etc.}}
                             , leaf]
                        ]
                        [\textbf{AI Feedback}, fill=align
                            [Constitutional AI~\citep{bai2022constitutional}{,}
                             RLAIF~\citep{lee2023rlaif}{,}
                             { \textit{etc.}}
                             , leaf]
                        ]
                    ]
                    [\textbf{New Methods4RM Design (\S\ref{sec3.3})}, fill=align
                        [\textbf{Principle/AI‑generated Rewards}, fill=align
                            [RewardAnything~\citep{yu2025rewardanything}{,}
                             AUTORULE~\citep{wang2025autorule}{,}
                             Generalist Reward Models~\citep{li2025generalist}{,}
                             { \textit{etc.}}
                             , leaf]
                        ]
                        [\textbf{Reasoning Reward Model}, fill=align
                            [RRMs~\citep{guo2025reward}{,}
                             GenPRM~\citep{zhao2025genprm}{,}
                             RM-R1~\citep{chen2025rm}{,}
                             Wang \textit{et al.}~\citep{wang2025unified}{,}
                             Liu \textit{et al.}~\citep{liu2025inference}{,}
                             { \textit{etc.}}
                             , leaf]
                        ]
                    ]
                ]
                [\textbf{RLVR (\S\ref{sec4})}, fill=rlvr
                    [\textbf{Algorithmic Advances (\S\ref{sec4.2})}, fill=rlvr
                        [\textbf{Type R1 and its improvements}, fill=rlvr
                            [GRPO~\citep{shao2024deepseekmath}{,}
                             Deepseek R1~\citep{guo2025deepseek}{,}
                             DAPO~\citep{yu2025dapo}{,}
                             Hu \textit{et al.}~\citep{hu2025open}{,}
                             X‑Reasoner~\citep{liu2025x}{,}\textbf{}\\
                             TANGO~\citep{zha2025rl}{,}
                             Zhou \textit{et al.}~\citep{zhou2025reinforcing}{,}
                             Kimina‑Prover~\citep{wang2025kimina}{,}
                             Magistral~\citep{rastogi2025magistral}{,}\\
                             MiniMax‑M1~\citep{chen2025minimax}{,}
                             GSPO~\citep{zheng2025group}{,}
                             { \textit{etc.}}
                             , leaf]
                        ]
                        [\textbf{Hybrid learning strategy}, fill=rlvr
                            [Zhang \textit{et al.}~\citep{zhang2025distill}{,}
                             Yan \textit{et al.}~\citep{yan2025learning}{,}
                             SuperRL~\citep{liu2025superrl}{,}
                             KDRL~\citep{xu2025kdrl}{,}
                             { \textit{etc.}}
                             , leaf]
                        ]
                        [\textbf{Adversarial/Structured/} \\ \textbf{Multi-agent}, fill=rlvr
                            [SPIRAL~\citep{liu2025spiral}{,}
                             R2‑Reasoner~\citep{shao2025route}{,}
                             Graph‑R1~\citep{luo2025graph}{,}
                             { \textit{etc.}}
                             , leaf]
                        ]
                        [\textbf{Tree structure}, fill=rlvr
                            [ToTRL~\citep{wu2025totrl}{,}
                             TreeRPO~\citep{yang2025treerpo}{,}
                             TreeRL~\citep{hou2025treerl}{,}
                             { \textit{etc.}}
                             , leaf]
                        ]
                    ]
                    [\textbf{Multimodal reasoning (\S\ref{sec4.3})}, fill=rlvr
                        [\textbf{Vision‑language reasoning}, fill=rlvr
                            [Vision‑R1~\citep{huang2025vision}{,}
                             Visual-RFT~\cite {liu2025visual}{,}
                             VLM‑R1~\citep{shen2025vlm}{,}
                             Deng \textit{et al.}~\citep{deng2025boosting}{,}
                             Ma \textit{et al.}~\citep{ma2025deepperception}{,}\\
                             Zhou \textit{et al.}~\citep{zhou2025r1}{,}\textbf{}
                             VisuLogic~\citep{xu2025visulogic}{,}
                             R1‑VL~\citep{zhang2025r1}{,}
                             GLM‑4.1V‑Thinking~\citep{hong2025glm}{,}
                             { \textit{etc.}}
                             , leaf]
                        ]
                        [\textbf{Video/Spatial and} \\\textbf{Embodied Reasoning}, fill=rlvr
                            [Video‑R1~\citep{feng2025video}{,}
                             TinyLLaVA‑Video‑R1~\citep{zhang2025tinyllava}{,}
                             3D‑R1~\citep{huang20253d}{,}
                             Liao \textit{et al.}~\citep{liao2025improved}{,}\textbf{}\\
                             Ouyang \textit{et al.}~\citep{ouyang2025spacer}{,}
                             Embodied‑R~\citep{zhao2025embodied}{,}
                             Ego‑R1~\citep{tian2025ego}{,}
                             VAU‑R1~\citep{zhu2025vau}{,}
                             { \textit{etc.}}
                             , leaf]
                        ]
                        [\textbf{o3 style}, fill=rlvr
                            [GPT o3~\citep{o3-o4-mini-system-card}{,}
                             DeepEyes~\citep{zheng2025deepeyes}{,}
                             GThinker~\citep{zhan2025gthinker}{,}
                             { \textit{etc.}}
                             , leaf]
                        ]
                        [\textbf{Generation and pure vision}, fill=rlvr
                            [T2I‑R1~\citep{jiang2025t2i}{,}
                             DanceGRPO~\citep{xue2025dancegrpo}{,}
                             Visual Planning~\citep{xu2025visual}{,}
                             { \textit{etc.}}
                             , leaf]
                        ]
                        [\textbf{Enhanced reasoning}, fill=rlvr
                            [Hint‑GRPO~\citep{huang2025boosting}{,}
                             Liu \textit{et al.}~\citep{liu2025more}{,}
                             SRPO~\citep{wan2025srpo}{,}
                             Wang \textit{et al.}~\citep{wang2025vicrit}{,}
                             VGR~\citep{wang2025vgr}{,}
                             { \textit{etc.}}
                             , leaf]
                        ]
                        [\textbf{Tasks in professional fields}, fill=rlvr
                            [VLN‑R1~\citep{qi2025vln}{,}
                             Chen \textit{et al.}~\citep{chen2025compile}{,}
                             VAU‑R1~\citep{zhu2025vau}{,}
                             ARMed~\citep{liu2025breaking}{,}
                             CAD‑Coder~\citep{guan2025cad}{,}
                             { \textit{etc.}}
                             , leaf]
                        ]
                    ]
                    [\textbf{Adaptive thinking (\S\ref{sec4.4})}, fill=rlvr
                        [\textbf{Length Penalty}, fill=rlvr
                            [S1~\citep{muennighoff2025s1}{,}
                             L1~\citep{aggarwal2025l1}{,}
                             Shorterbetter~\citep{yi2025shorterbetter}{,}
                             { \textit{etc.}}
                             , leaf]
                        ]
                        [\textbf{Thinking Mode} \& \\\textbf{Difficulty‑awareness}, fill=rlvr
                            [AdaptThink~\citep{zhang2025adaptthink}{,}
                             AdaCoT~\citep{chen2025rm}{,}
                             Zhang \textit{et al.}~\citep{zhang2025continue}{,}
                             Wang \textit{et al.}~\citep{wang2025adaptive}{,}
                             { \textit{etc.}}
                             , leaf]
                        ]
                        [\textbf{GRPO‑Variant}, fill=rlvr
                            [RRMs~\citep{guo2025reward}{,}
                             DeGRPO~\citep{fang2025thinkless}{,}
                             HGPO~\citep{jiang2025think}{,}
                             Wang \textit{et al.}~\citep{wang2025unified}{,}
                             { \textit{etc.}}
                             , leaf]
                        ]
                    ]
                    [\textbf{Agents/Tool‑use (\S\ref{sec4.5})}, fill=rlvr
                        [\textbf{Tool/Action Space RLVR}, fill=rlvr
                            [AGILE~\citep{peiyuan2024agile}{,}
                             Search-R1~\cite{jin2025search}{,}
                             ToRL~\citep{li2025torl}{,}
                             ReTool~\citep{feng2025retool}{,}
                             OpenThinkImg~\citep{su2025openthinkimg}{,}\\
                             Tool‑Star~\citep{dong2025tool}{,}
                             { \textit{etc.}}
                             , leaf]
                        ]
                        [\textbf{Long‑term Reward Design}, fill=rlvr
                            [LARM~\citep{li2024larm}{,}
                             RAGEN~\citep{wang2025ragen}{,}
                             ShopR1~\citep{zhang2025shopr1rewardingllmssimulate}{,}
                             Zhang \textit{et al.}~\citep{zhang2025divide}{,}
                             SPA‑RL~\citep{wang2025spa}{,}\\
                             Feng \textit{et al.}~\citep{feng2025group}{,}
                             { \textit{etc.}}
                             , leaf]
                        ]
                        [\textbf{Agent Memory}, fill=rlvr
                            [MemAgent~\cite{yu2025memagent}{,}
                             RMM~\cite{tan2025prospect}{,}
                             M3-Agent~\cite{long2025seeing}{,}
                             Memory-R1~\cite{yan2025memory}{,}
                             { \textit{etc.}}
                             , leaf]
                        ]
                    ]
                    [\textbf{RLIF/Internal Feedback (\S\ref{sec4.6})}, fill=rlvr
                        [\textbf{Internal Signal Reward}, fill=rlvr
                            [Kang \textit{et al.}~\citep{kang2025scalable}{,}
                             Li \textit{et al.}~\citep{li2025confidence}{,}
                             van Niekerk \textit{et al.}~\citep{vanniekerk2025posttraininglargelanguagemodels}{,}
                             Zhang \textit{et al.}~\citep{zhang2025no}{,}
                             { \textit{etc.}}
                             , leaf]
                        ]
                        [\textbf{Self‑generation Optimization}, fill=rlvr
                            [TTRL~\citep{zuo2025ttrl}{,}
                             Zhao \textit{et al.}~\citep{zhao2025absolute}{,}
                             Zhao \textit{et al.}~\citep{zhao2025learning}{,}
                             SLOT~\citep{hu2025slot}{,}
                             { \textit{etc.}}
                             , leaf]
                        ]
                    ]
                    [\textbf{Experimental Findings (\S\ref{sec4.1})}, fill=rlvr
                        [\textbf{Capability Boundaries}, fill=rlvr
                            [Yue \textit{et al.}~\citep{yue2025does}{,}
                             ProRL~\citep{liu2025prorl}{,}
                             Liu \textit{et al.}~\citep{liu2025there}{,}
                             Zhao \textit{et al.}~\citep{zhao2025echo}{,}
                             Shah \textit{et al.}~\citep{shah2025rethinking}{,}\\
                             Wu \textit{et al.}~\citep{wu2025invisible}{,}
                             { \textit{etc.}}
                             , leaf]
                        ]
                        [\textbf{Training Dynamics}\\ \textbf{\& Entropy Signals}, fill=rlvr
                            [Cui \textit{et al.}~\citep{cui2025entropy}{,}
                             Li \textit{et al.}~\citep{li2025think}{,}
                             Ma \textit{et al.}~\citep{ma2025reasoning}{,}
                             Samineni \textit{et al.}~\citep{samineni2025rl}{,}
                             Bogdan \textit{et al.}~\citep{bogdan2025thought}{,}\\
                             He \textit{et al.}~\citep{he2025response}{,}
                             Chen \textit{et al.}~\citep{chen2025sft}{,}
                             Wang \textit{et al.}~\citep{wang2025beyond}{ \textit{etc.}}
                             , leaf]
                        ]
                        [\textbf{Data \& Reward Conditions}, fill=rlvr
                            [Wu \textit{et al.}~\citep{wu2025reasoning}{,}
                             Zhu \textit{et al.}~\citep{zhu2025surprising}{,}
                             Shojaee \textit{et al.}~\citep{shojaee2025illusion}{,}
                             { \textit{etc.}}
                             , leaf]
                        ]
                    ]
                ]
                [\textbf{Datasets \& Benchmarks (\S\ref{sec5})}, fill=db
                    [\textbf{Synthetic Data Generation}, fill=db
                        [Zhu \textit{et al.}~\citep{askell2021general}{,}
                         Goldie \textit{et al.}~\citep{goldie2025synthetic}{,}
                         Synthetic Data RL~\citep{guo2025synthetic}{,}
                         SwS~\citep{liang2025sws}{,}
                         { \textit{etc.}}
                         , leaf]
                    ]
                    [\textbf{Alignment/Dialogue}, fill=db
                        [HHH~\citep{askell2021general}{,}
                         HH‑RLHF~\citep{bai2022training}{,}
                         IFEval\cite{zhou2023instruction}{,}
                         Arena-Hard\cite{li2024crowdsourced}{,}
                         AlignBench\cite{liu-etal-2024-alignbench}{,}\\
                         Creative Writing\cite{creative-writing-bench-v3}{,}
                         { \textit{etc.}}
                         , leaf]
                    ]
                    [\textbf{Code}, fill=db
                        [APPS~\citep{dou2024stepcoder}{,}
                         LiveCodeBench~\citep{jain2024livecodebench}{,}
                         SWE‑bench~\citep{lewkowycz2022solving}{,}
                         SWE‑bench Verified~\citep{yang2025swe}{,}\\
                         OJBench~\citep{wang2025ojbench}{,}
                         { \textit{etc.}}
                         , leaf]
                    ]
                    [\textbf{Math}, fill=db
                        [GSM8K~\citep{cobbe2021training}{,}
                         MATH~\citep{hendrycks2021measuring}{,}
                         OlympiadBench~\citep{he2024olympiadbench}{,}
                         Minerva Math~\citep{lewkowycz2022solving}{,}
                         PolyMath~\citep{wang2025polymath}{,}\\
                         AMC2023{,} AIME2024/2025{,} CNMO2024{,} HMMT2025{,} { \textit{etc.}}
                         , leaf]
                    ]
                    [\textbf{General / Knowledge \& STEM}, fill=db
                        [MMLU~\citep{hendrycks2020measuring}{,}
                         MMLU-Redux\cite{gema2024we}{,}
                         MMLU‑Pro~\citep{wang2024mmlu}{,}
                         GPQA~\citep{rein2024gpqa}{,}
                         SuperGPQA~\citep{du2025supergpqa}{,}\\
                         TheoremQA~\citep{chen2023theoremqa}{,}
                         Guru~\citep{cheng2025revisiting}{,}
                         SimpleQA~\citep{wei2024measuring}{,}
                         HLE~\citep{phan2025humanity}{,}
                         LiveBench~\citep{White2025LiveBench}{,}
                         PhyX~\citep{shen2025phyx}{,}\\
                         BBH~\citep{srivastava2023beyond}{,}
                         BBEH~\citep{kazemi2025big}{,}
                         MMReason~\citep{yao2025mmreason}{,}
                         { \textit{etc.}}
                         , leaf]
                    ]
                    [\textbf{Logic Reasoning}, fill=db
                        [AutoLogi~\citep{zhu2025autologi}{,}
                         ZebraLogic~\citep{lin2025zebralogic}{,}
                         { \textit{etc.}}
                         , leaf]
                    ]
                    [\textbf{Tools / Multi‑turn / Agent}, fill=db
                        [$\tau^2$‑Bench~\citep{barres2025tau}{,}
                         ACEBench~\citep{chen2025acebench}{,}
                         MultiChallenge~\citep{sirdeshmukh2025multichallenge}{,}
                         { \textit{etc.}}
                         , leaf]
                    ]
                ]
                [\textbf{Open‑source Tools}\\\textbf{\&Frameworks (\S\ref{sec6})}, fill=tool
                    [\textbf{General \& End‑to‑End}\\ \textbf{Frameworks}, fill=tool
                        [VeRL~\citep{sheng2025hybridflow}{,}
                         ColossalChat~\citep{you2023colossalchat}{,}
                         DeepSpeed‑Chat~\citep{yao2023deepspeed}{,}
                         TRL~\citep{vonwerra2022trl}{,}
                         RL4LMs~\citep{ramamurthy2022reinforcement}{,}\\
                         LlamaRL~\citep{wu2025llamarl}{,}
                         trlX~\citep{havrilla2023trlx}{,}
                         AReaL~\citep{fu2025areal}{,}
                         DistFlow~\citep{wang2025distflow}
                         OpenRLHF~\citep{hu2024openrlhf}{,}
                         { \textit{etc.}}
                         , leaf]
                    ]
                    [\textbf{RL Training Library}\\\textbf{/Package}, fill=tool
                        [Nemo RL~\citep{nemo-rl}{,}
                         FlashRL~\citep{liu2025flashrl}{,}
                         ROLL~\citep{wang2025reinforcement}{,}
                         Yao \textit{et al.}~\citep{yao2025offpolicy}{,}
                         { \textit{etc.}}
                         , leaf]
                    ]
                ]
            ]
        \end{forest}
    }
    \caption{A taxonomy of RL enhanced LLMs. This figure presents a taxonomy of the key stages and resources involved in creating RL-enhanced LLMs, organized into five branches: pre-training, alignment, RLVR, datasets \& benchmarks, and open-source frameworks. The taxonomy clarifies the interconnections between stages, serving as a roadmap for understanding methodological advancements and resources discussed in the survey.}
    \vspace{-0.4cm}
    \label{fig:rl_overview}
\end{figure*}
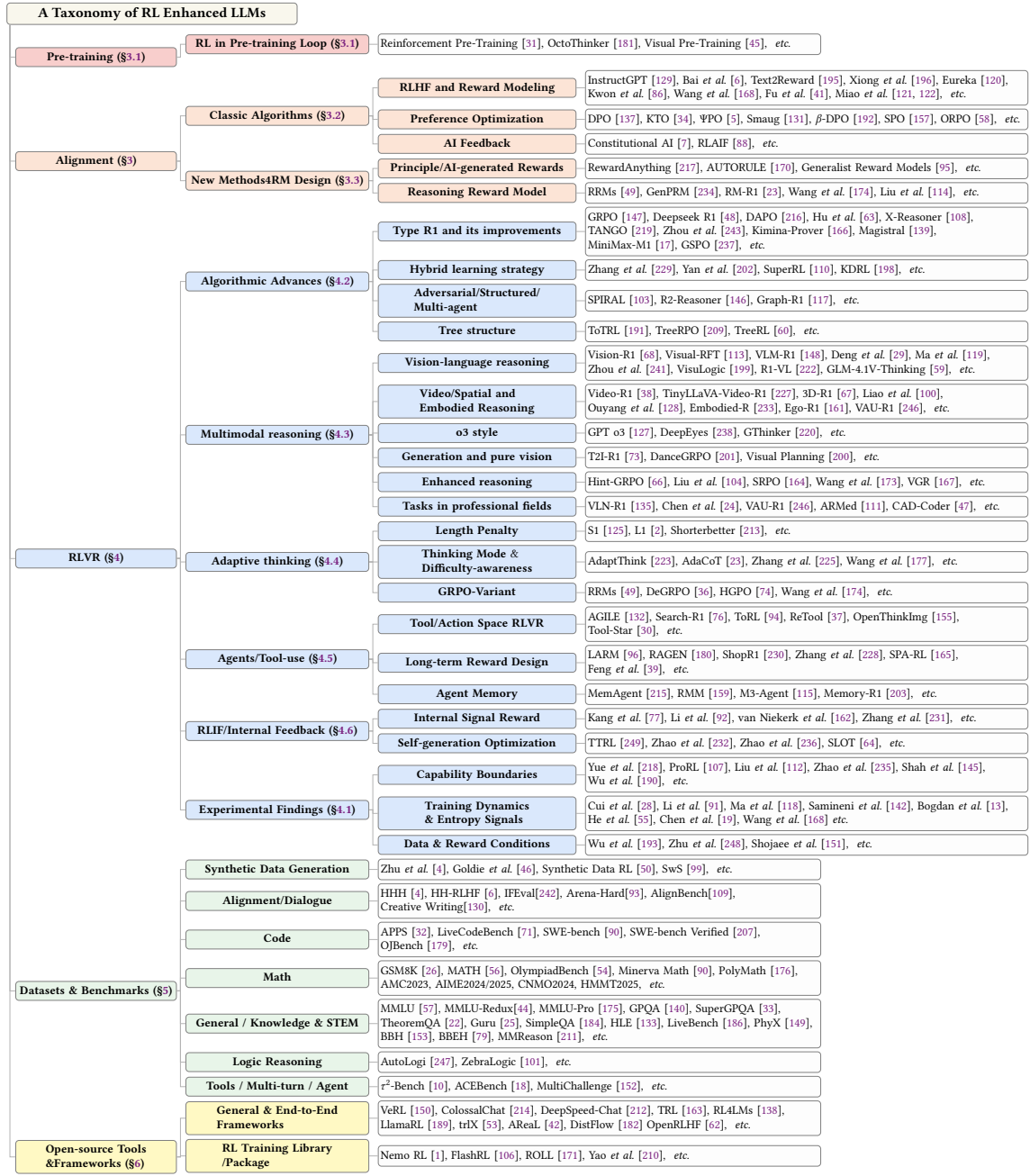

In light of these gaps, this survey aims to provide a systematic and comprehensive review of recent progress in RL-enhanced LLMs, with particular focus on developments in the highly influential RLVR paradigm, especially with rapid developments since 2025. We aim to clarify the role of RL methods in the entire LLM training pipeline and their contributions to advancing the frontiers of model alignment and reasoning. Specifically, we offer in-depth analysis and discussion along multiple dimensions: (1) the theoretical foundations of applying RL to LLMs; (2) application strategies detailing how RL is integrated at different training stages, including initial pre-training, alignment fine-tuning, and post-training inference-time reasoning; (3) the datasets and benchmarks used to train and evaluate RL-fine-tuned LLMs; and (4) the emerging tools and frameworks that support large-scale RL training for LLMs. By organizing the survey along these axes, we aim to provide researchers and practitioners with a clear roadmap of the field’s current state, insights into the efficacy and limitations of various RL techniques especially RLVR, and well-supported guidance for future work in leveraging RL to make LLMs more aligned, powerful, and reliable.

\subsection{Related Surveys}
\newcommand{\cmark}{\textcolor{green!60!black}{\checkmark}}
\newcommand{\xmark}{\textcolor{red}{\times}}
\newcolumntype{C}[1]{>{\centering\arraybackslash}p{#1}}

\begin{table}[tbp]
\centering
\caption{Comparative Analysis Table of Representative Surveys: The comparison is conducted across five dimensions—lifecycle coverage, dataset and benchmark summarization, tool/framework collection and practicality, breadth and timeliness of citations, and future outlook and challenges.}
\vspace{1mm}
\label{tab:survey_comparison}
\setlength{\tabcolsep}{4pt} 
\renewcommand{\arraystretch}{1.4}
\footnotesize
\begin{tabularx}{\textwidth}{l p{2.2cm} p{2.2cm} p{2.2cm} p{2.2cm} p{2.2cm}}
\toprule
Survey $\downarrow$  / Dimension $\rightarrow$ &
Lifecycle Coverage &
Datasets\&Benchmarks &
Tools / Frameworks &
\shortstack{Citation Breadth \\ \& Timeliness} &
\shortstack{Future Directions\\ \& Challenges} \\
\midrule
Wang \textit{et al.}~\cite{wang2024comprehensive} &
$\xmark$ (Alignment only) & 
$\xmark$ (Limited mention) &
$\xmark$ (Not covered) &
$\xmark$ (Insufficiently updated) &
$\cmark$ (Future directions mentioned) \\
Srivastava \textit{et al.}~\cite{srivastava2025technical} &
$\xmark$ (Alignment + Reasoning) &
$\xmark$ (For demonstrating performance only) &
$\xmark$ (Not covered) &
$\cmark$ (Covers up to 2025) &
$\cmark$ (Dedicated section) \\
Wang \textit{et al.}~\cite{wang2024reinforcement} &
$\xmark$ (Mainly Alignment) &
$\xmark$ (Limited mention) &
$\xmark$ (Not covered) &
$\cmark$ (Covers early 2025) &
$\cmark$ (Simpler discussion) \\
Cao \textit{et al.}~\cite{cao2024survey} &
$\xmark$ (Not all covered) &
$\xmark$ (Not covered) &
$\xmark$ (Not covered) &
$\xmark$ (Not included) &
$\cmark$ (Dedicated section) \\
Chaudhari \textit{et al.}~\cite{chaudhari2024rlhf} &
$\xmark$ (Alignment only) &
$\xmark$ (Limited mention) &
$\xmark$ (Not covered) &
$\xmark$ (Focus on RLHF only, outdated) &
$\cmark$ (In-depth analysis) \\
Kaufmann \textit{et al.}~\cite{kaufmann2024surveyreinforcementlearninghuman} &
$\xmark$ (Alignment only) &
$\cmark$ (Dedicated section) &
$\cmark$ (Library support mentioned) &
$\xmark$ (Relatively early) &
$\cmark$ (Brief analysis) \\
\textbf{Our Survey} &
$\cmark$ (Full coverage) &
$\cmark$ (Dedicated section) &
$\cmark$ (Well organized) &
$\cmark$ (Covers latest work) &
$\cmark$ (In-depth analysis) \\
\bottomrule
\end{tabularx}
\end{table}

In recent years, numerous surveys~\cite{cao2024survey, wang2024reinforcement, pternea2024rl, srivastava2025technical, wang2024comprehensive, zhou2025reinforced, bandyopadhyay2025thinking, ke2025survey, ji2025survey, xu2025towards, chaudhari2024rlhf, jiang2024survey, hao2025rl, kumar2025llm, zhong2025comprehensive, kaufmann2024surveyreinforcementlearninghuman, besta2025reasoning, zhu2025towards, zhang2025survey} have reviewed reinforcement learning research related to large language models and proposed various classification schemes. Existing surveys have proposed a variety of classification schemes, but often with a limited scope. For example, some studies~\cite{kaufmann2024surveyreinforcementlearninghuman, wang2024comprehensive, zhong2025comprehensive} narrowly focus only on RL-based alignment techniques, organizing their taxonomies primarily around the use of reward models while overlooking important emerging approaches. Although several works in 2025 have attempted to summarize research on RL at inference time~\cite{xu2025towards, besta2025reasoning, bandyopadhyay2025thinking, ke2025survey, zhou2025reinforced}, these reviews are often partial and fail to provide a holistic examination of reinforcement-at-inference across its multiple dimensions. Pternea \textit{et al.}~\cite{pternea2024rl} discuss the synergy between RL and LLMs, but their analysis is largely limited to the perspective of bidirectional RL–LLM collaboration. Zhu \textit{et al.}~\cite{zhu2025towards} focuses exclusively on the narrow domain of Concise and Adaptive Thinking. While these survey frameworks offer value, they remain constrained to specific viewpoints and lack a unified, end-to-end lifecycle perspective on RL–LLM interactions. In contrast, our survey systematically investigates the role of RL throughout the entire LLM training pipeline (ranging from pre-training and alignment fine-tuning reasoning) and proposes an organizational framework that, to the best of our knowledge, has not been comprehensively addressed in prior research. Table \ref{tab:survey_comparison} summarizes the advantages and disadvantages of our survey compared with other representative surveys.

\subsection{Contribution Summary}
\begin{figure}[!htbp]
  \centering
  \includegraphics[width=0.9\linewidth]{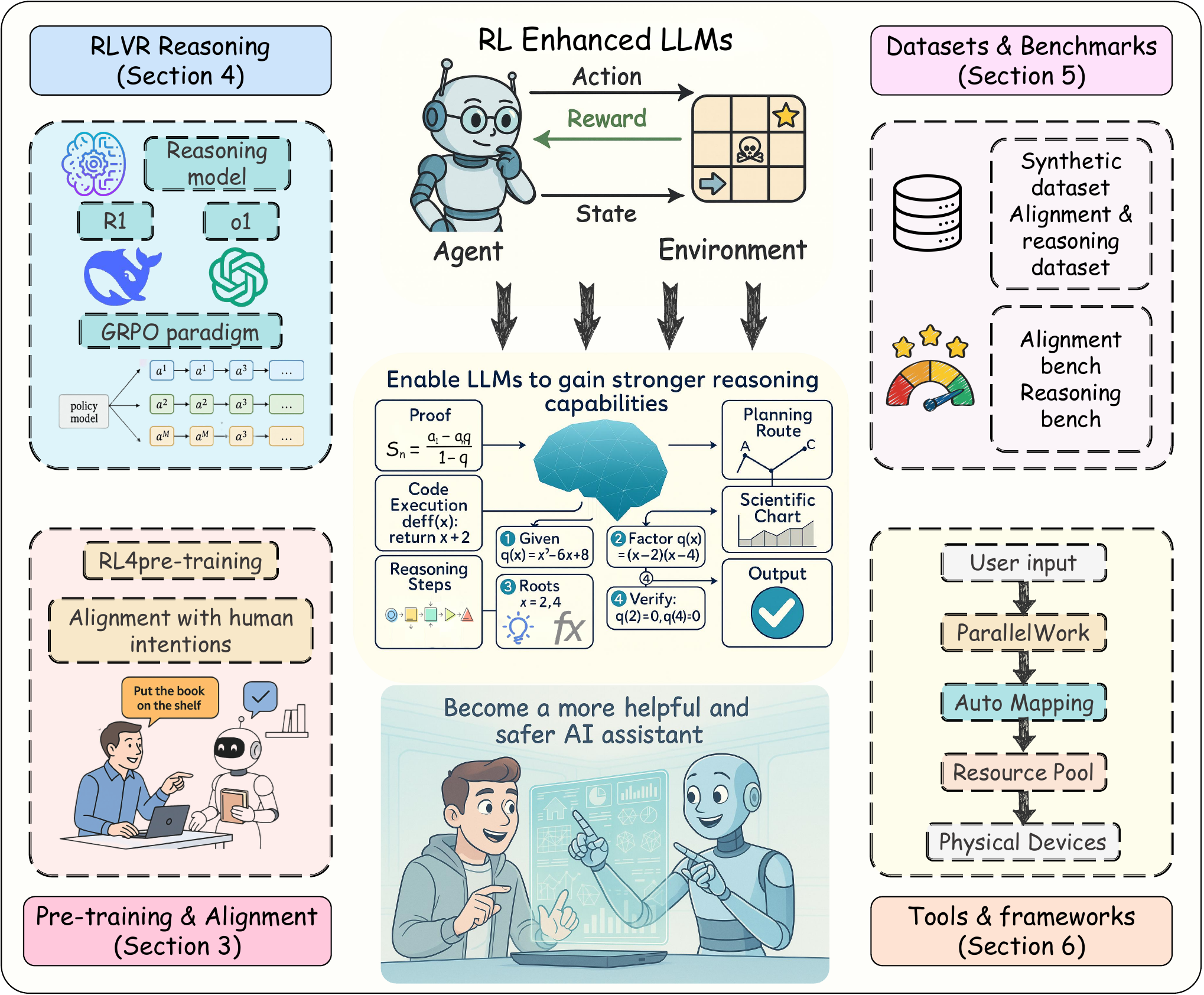 }
  \caption{Key components in RL-enhanced LLMs. This figure illustrates the key components and their interactions within the lifecycle of RL-enhanced LLMs. Driven by RL frameworks and toolkits, RL algorithms participate in the pre-training, alignment, and reasoning enhancement training of LLMs, and are validated through test benchmarks.}
  \label{zongtu}
  \vspace{-7pt}
\end{figure}

This survey provides a structured review of RL techniques for LLMs, with three distinctive contributions:

\begin{itemize}
    \item \textbf{Lifecycle Organization:} We systematically cover the full lifecycle of RL for LLMs, detailing each stage of the process, from pre-training, alignment, to reinforcement for reasoning. In doing so, we clarify the objectives, methodologies, and challenges encountered at each phase. This organization helps in understanding how RL techniques are applied and refined throughout the LLM development lifecycle.
    
    \item \textbf{Advanced RLVR Technology Focus:} This paper highlights state-of-the-art approaches in RLVR. We provide an in-depth analysis of the experimental phenomena and cutting-edge applications of RLVR, exploring the methodologies used to ensure that rewards are objective and verifiable. Additionally, we discuss how verifiable rewards contribute to improved model performance and alignment, showcasing the strengths and limitations of RLVR in real-world applications.
    
    \item \textbf{Consolidated Resources:} We summarize the datasets, benchmarks, and open-source frameworks that are critical for RL-based experimentation, evaluation, and practical implementation in LLMs. By aggregating this information, we provide a valuable resource for future researchers looking to experiment with RL techniques in the context of LLMs. The inclusion of these resources enhances the reproducibility and transparency of RL-driven LLM research.
\end{itemize}
To provide an organizational roadmap, Figure \ref{fig:rl_overview}  presents a comprehensive taxonomy, which divides existing approaches into five branches: pre-training, alignment, RLVR, datasets \& benchmarks, and open-source frameworks. As outlined in Figure \ref{zongtu}, our review is organized around the full RL lifecycle for LLMs, with a particular emphasis on RL with verifiable rewards. In summary, this survey delivers a lifecycle-based synthesis of methods, with particular emphasis on RLVR, complemented by practical resources for research and application.

\section{Preliminaries  of Reinforcement Learning}~\label{sec2}
Reinforcement learning enables agents to learn optimal policies through interaction with the environment, aiming to maximize cumulative rewards. A typical RL problem can be modelled as a Markov Decision Process (MDP), which consists of a state space, an action space, a state transition probability distribution, and a reward function. At each timestep, the agent selects an action $a$ based on the current state $s$, receives an immediate reward $r$, and transitions to a new state $s'$ according to the environment dynamics. The objective of the agent is to learn an optimal policy $\pi^*$ that maximizes the expected long-term cumulative reward over the course of interactions. To achieve this objective, RL algorithms have evolved along two primary paradigms: policy-based and value-based learning. The former directly focuses on optimizing the policy, often through policy gradient methods; the latter emphasizes estimating the value of states or actions, from which the policy is derived indirectly. This section introduces representative algorithms and theoretical foundations of these two RL paradigms, and further discusses their applications in LLM training.
\subsection{Policy Learning}
Policy learning methods directly optimize the policy $\pi(a|s;\theta)$, typically without explicitly learning an environment model or value function. A common approach is the policy gradient method, which adjusts policy parameters $\theta$ in the parameter space through gradient ascent to maximize expected returns. 
REINFORCE~\cite{williams1992simple} is the most fundamental Monte Carlo policy gradient method. It directly estimates the gradient of the expected return with respect to the policy parameters, where the objective is defined as $J(\theta) = \mathbb{E}[R]$, with $R$ denoting the cumulative return. By applying the log-derivative trick for stochastic policies, an unbiased estimator of the policy gradient can be derived as:
\begin{equation}
\nabla_\theta J(\theta) = \mathbb{E}_{\tau \sim \pi_\theta}\left[\sum_{t=0}^T \nabla_\theta \log \pi_\theta(a_t | s_t) R_t\right].
\end{equation}
Here, $\tau = (s_0, a_0, r_0, \dots, s_T)$ denotes a trajectory, and $R_t = \sum_{k=t}^{T} \gamma^{k-t} r_k$ is the discounted return starting from timestep $t$, where $\gamma$ is the discount factor. Intuitively, this formulation implies that increasing the probability of taking action $a_t$ in state $s_t$ is positively correlated with the cumulative reward obtained thereafter~\cite{sutton1999policy}. After sampling a complete trajectory, REINFORCE updates the policy parameters using the estimated gradient as $\theta \leftarrow \theta + \alpha \nabla_\theta J(\theta)$, where $\alpha$ is the learning rate. To reduce the variance of the gradient estimator, a common technique is to introduce a baseline function $b(s)$, which depends only on the state. Subtracting this baseline from the return does not bias the gradient estimate but can significantly reduce its variance. The policy gradient estimator with a baseline is thus written as:
\begin{equation}
\nabla_\theta J(\theta)=\mathbb{E}\left[\sum_t\nabla_\theta\log\pi_\theta(a_t|s_t)\left(R_t-b(s_t)\right)\right].
\end{equation}
$A_t = R_t - b(s_t)$  is the advantage function, which measures how much the actual return exceeds the baseline, reducing the variance of the gradient estimator without altering its expectation.

The Actor-Critic (AC) method~\cite{konda1999actor} combines policy gradient techniques with value function approximation by integrating two components within a unified framework: the actor, which selects actions according to a parameterized policy $\pi_\theta(a|s)$, and the critic, which evaluates the policy using a parameterized value function $V_\phi(s)$ or action-value function $Q_\phi(s, a)$. At each timestep, the AC algorithm alternates between: (1) the critic estimating the advantage $A(s, a) = Q(s, a) - V(s)$ or the Temporal-Difference (TD) error $\delta = r + \gamma V(s') - V(s)$; and (2) the actor updating the policy parameters along the direction of the policy gradient, weighted by the low-variance estimate provided by the critic, \textit{i.e.}, using $\nabla_\theta \log \pi_\theta(a|s) \cdot A(s, a)$ as the gradient.

Trust Region Policy Optimization (TRPO)~\cite{schulman2015trust} aims to address the instability that can arise from large policy updates. TRPO formulates policy optimization as a constrained optimization problem: it maximizes the expected advantage under the old policy $\pi_{\text{old}}$, while constraining the KL divergence between the new and old policies to remain below a threshold $\delta$. The formal objective is given by:
\begin{equation}
\begin{aligned}\max_\theta&L(\theta)=\mathbb{E}_{s\sim\pi_{\mathrm{old}}}\left[\sum_a\frac{\pi_\theta(a|s)}{\pi_{\mathrm{old}}(a|s)}A^{\pi_{\mathrm{old}}}(s,a)\right],\\\\\mathrm{s.t.~}&\mathbb{E}_{s\sim\pi_{\mathrm{old}}}\left[D_{\mathrm{KL}}(\pi_{\mathrm{old}}(\cdot|s)\parallel\pi_\theta(\cdot|s))\right]\leq\delta.\end{aligned}
\end{equation}
Proximal Policy Optimization (PPO) ~\cite{schulman2017proximal} is a landmark innovation of traditional policy gradient algorithms in the era of deep reinforcement learning. The core contribution is a clipped surrogate objective that allows multi-step gradient updates without policy collapse. Specifically, PPO uses the objective function:
\begin{equation}
    L^{\mathrm{PPO}}(\theta)=\mathbb{E}_t[\min(r_t(\theta)\hat{A}_t,\mathrm{~clip}(r_t(\theta),1-\epsilon,1+\epsilon)\hat{A}_t)].
\end{equation}
Here, $r_t(\theta)$ represents the probability ratio of the old and new policies on the action at time $t$, $\epsilon$ is the threshold, and $\hat{A}_t$ is the advantage estimate. This objective enables optimization according to the standard policy gradient when the magnitude of the policy change is within the threshold; once it exceeds the threshold, the gradient is weakened, thereby ensuring that a single update will not cause the policy to deviate too far from the original policy.

In LLM fine-tuning, PPO optimizes parameters via reward scores with a value network baseline for efficient advantage estimation, but despite its stability and success in RLHF alignment, reasoning tasks face added memory/computation costs from the extra value network and instability in long sequences due to inaccurate value estimates. To address this issue, as well as the low learning efficiency in traditional RLHF settings where only one response is scored at a time, the DeepSeek team proposed Group Relative Policy Optimization (GRPO) in DeepSeekMath~\cite{shao2024deepseekmath}. The core idea of GRPO is to sample a set of outputs for each prompt and use the relative differences in intra-group feedback to guide policy updates. Specifically, for each question-answer pair, the behavioral policy of GRPO generates $G$ different answers at once (forming a group). Then, each answer is assigned a reward value $R_i$ through a reward model or predefined rules. Instead of training a separate value function to estimate a global baseline, GRPO adopted the intra-group average reward as the benchmark: it calculates the average or a certain statistic of the rewards of all answers in the group, and defines the advantage of each answer as $A_i = R_i - \bar{R}_{\text{group}}$. In this way, answers with rewards higher than the group average gain positive advantages, while those lower than the average gain negative advantages. Subsequently, GRPO constructs a clipped policy objective function similar to PPO, and through gradient ascent, it increases the probability of answers with positive advantages and decreases the probability of answers with negative advantages. Since the group average serves as a dynamic baseline, advantages can be calculated without additional training of a value network, thereby simplifying the algorithm structure. For a specific question-answer pair $(q, a)$, the behavioral policy $\pi_{\theta_{\mathrm{old}}}$ samples and generates $G$ independent responses $\{o_i\}_{i=1}^G$. Subsequently, the advantage value of the i-th response is calculated by normalizing the group-level rewards:  
\begin{equation}
\hat{A}_{i,t}=\frac{r_i-\max(\{R_i\}_{i=1}^G)}{\operatorname{std}(\{R_i\}_{i=1}^G)}.\end{equation}
Similar to PPO, GRPO employs a clipped objective function and directly introduces a KL penalty term. The objective function of GRPO is as follows:
\begin{equation}
\begin{aligned}\mathcal{J}_{\mathrm{GRPO}}(\theta)&=\mathbb{E}_{(q,a)\sim\mathcal{D},\{o_i\}_{i=1}^G\sim\pi_{\theta_{\mathrm{old}}}(\cdot|q)}\\&\left[\frac{1}{G}\sum_{i=1}^G\frac{1}{|o_i|}\sum_{t=1}^{|o_i|}\left(\min\left(r_{i,t}(\theta)\hat{A}_{i,t},\right.\operatorname{clip}\left(r_{i,t}(\theta),1-\varepsilon,1+\varepsilon\right)\hat{A}_{i,t}\right)-\beta D_{\operatorname{KL}}(\pi_{\boldsymbol{\theta}}||\pi_{\operatorname{ref}})\right).\end{aligned}
\end{equation}
\subsection{Value Learning}
Value-based methods aim to indirectly derive the optimal policy by estimating value functions. A value function quantifies the expected long-term utility of a state or state-action pair under a given policy. Typical examples include the state-value function $V^\pi(s) = \mathbb{E}_\pi[R \mid s]$ and the action-value function $Q^\pi(s, a) = \mathbb{E}_\pi[R \mid s, a]$. Value-based algorithms focus on approximating the optimal value function $V(s)$ or $Q(s, a)$, and then deriving the optimal policy by following the value maximization principle—\textit{e.g.}, selecting the action with the highest estimated value at each state.

Q-learning~\cite{watkins1992q} adopted a model-free, off-policy learning approach to approximate the optimal action-value function $Q^*(s, a)$. The core idea is to iteratively update value estimates for state-action pairs, guided by the Bellman optimality equation. The basic Q-learning update rule is given by:
\begin{equation}
Q_{new}(s_t,a_t)\leftarrow Q(s_t,a_t)+\alpha\begin{bmatrix}r_t+\gamma\max_{a^{\prime}}Q(s_{t+1},a^{\prime})-Q(s_t,a_t)\end{bmatrix},
\end{equation}
where $\alpha$ denotes the learning rate, and $\gamma$ is the discount factor. The update rule corrects the current estimate $Q(s_t, a_t)$ by incorporating the temporal-difference (TD) error: $\delta = r + \gamma \max_{a'} Q(s_{t+1}, a') - Q(s_t, a_t)$, which reflects the discrepancy between the newly estimated return and the previous estimate of $Q(s_t, a_t)$. In each step, Q-learning collects experience tuples $(s_t, a_t, r_t, s_{t+1})$ using exploration strategies such as $\epsilon$-greedy, and then updates the corresponding $Q$ value. Under suitable conditions, the $Q(s, a)$ values converge to the true optimal action-value function $Q^*(s, a)$ in the long run. Since each update relies on the estimated maximum reward at the next state, \textit{i.e.}, $\max_{a'} Q(s_{t+1}, a')$, rather than the action actually taken, Q-learning is categorized as an off-policy method. This allows it to learn from historical data or samples generated by different policies. However, this also introduces an overestimation bias: the maximization step can yield upward-biased value estimates. In practice, several improvements have been proposed, such as Double Q-learning~\cite{hasselt2010double}, which mitigates overestimation by maintaining two separate value estimators.

SARSA~\cite{rummery1994line} is another temporal-difference-based value learning method. In contrast to Q-learning, SARSA is an on-policy algorithm: it evaluates the action values according to the currently executed policy and updates the estimates using samples collected from the same policy. The SARSA update rule is given by:
\begin{equation}
Q_{new}(s_t,a_t)\leftarrow Q(s_t,a_t)+\alpha\begin{bmatrix}r_t+\gamma Q(s_{t+1},a_{t+1})-Q(s_t,a_t)\end{bmatrix}.
\end{equation}
Unlike Q-learning, which uses the maximal action at the next state $s_{t+1}$, SARSA relies on the action $a_{t+1}$ actually taken by the agent according to the current policy (\textit{e.g.}, $\epsilon$-greedy). This implies that SARSA updates the estimate of $Q^\pi(s, a)$ under the current policy $\pi$. As the policy gradually improves toward a greedy strategy, the SARSA estimates of $Q$ progressively approach the optimal $Q^*$.

Deep Q-Network (DQN)~\cite{mnih2015human} represents a breakthrough in value-based methods by introducing neural function approximation into Q-learning. The core idea of DQN is to use a deep neural network $Q(s, a; \theta)$, parameterized by $\theta$, to approximate the action-value function. By adjusting the network parameters, $Q(s, a; \theta)$ outputs value estimates for all possible actions given a state input. DQN adopted the Q-learning target to train this network, aiming to make the predicted $Q$-values satisfy the Bellman optimality equation. Specifically, given a transition $(s, a, r, s')$ sampled from the experience replay buffer, DQN minimizes the difference between the predicted Q-value and the TD target using the mean squared error loss:
\begin{equation}
L(\theta) = \left(r + \gamma \max_{a'} Q(s', a'; \theta^-) - Q(s, a; \theta)\right)^2.
\end{equation}
Here, $\theta^{-}$ denotes the parameters of the target network, which is periodically synchronized from the training network parameters $\theta$, but kept fixed between updates. This dual-network mechanism enhances training stability by preventing divergence caused by simultaneous changes in both the target and prediction values.

In the domain of LLMs, value-based methods are not the primary components of training frameworks such as RLHF. This is largely due to the immense and complex action space in LLMs, 
making it infeasible to explicitly construct Q-tables or networks that evaluate the value of every possible output, as in standard reinforcement learning environments. Nonetheless, the conceptual foundations of value learning still manifest in LLM reinforcement learning. For instance, Wang \textit{et al.}~\cite{wang2024demonstration} adopted a Q-learning-based framework to dynamically select in-context exemplars. By computing a diversity score over label distributions among selected demonstrations, the framework jointly maximizes both diversity and task relevance, effectively guiding LLMs to generate more informative references for text classification.
\section{Reinforcement Learning Methods in the Pre-training Phase and Alignment Phase}~\label{sec3}
\subsection{Reinforcement Learning Methods in the Pre-training Phase}~\label{sec3.1}
Most current reinforcement learning enabled tasks for LLMs are mainly focused on the alignment and fine-tuning phases of the models. However, Dong \textit{et al.}~\cite{dong2025reinforcement} reconstructed the next-token prediction task in pre-training into an RL-based reasoning task, allowing the model to obtain verifiable rewards when correctly predicting the next token in a given context, thereby introducing reinforcement learning into pre-training tasks. Nevertheless, this method consumes excessive resources and requires an excellent model with existing reasoning capabilities as the base model for training. Ghosh \textit{et al.}~\cite{ghosh2025visual} introduced RL into visual pre-training with Annotation Bootstrapping, framing unlabeled image pre-training as an RL problem and arguing that common self-supervised methods like crop-consistency resemble value learning. OctoThinker~\cite{wang2025octothinker} proposed to significantly improve the compatibility of basic language models with reinforcement learning through a two-stage mid-training strategy, enabling the Llama series, which was originally unsuitable for RL, to reach the same level as Qwen in mathematical reasoning tasks. It reveals the key role of mid-training data quality, style, and scheduling strategies in RL scaling. Mid-training refers to self-supervised training conducted in the same way as pre-training, \textit{i.e.}, through next-word prediction, but with a different goal. The goal of mid-training is to transform the pre-trained model to make it suitable for RL training, and the training data is converted from massive amounts of text to high-quality, task-related data. 
\subsection{Classic Algorithms in the Alignment Phase}~\label{sec3.2}
Christiano \textit{et al.}~\cite{ouyang2022training} established a foundational paradigm for modern LLM alignment, demonstrating that incorporating human preferences into fine-tuning significantly improves the helpfulness and safety of instruction following behaviour. Bai \textit{et al.}~\cite{bai2022training} found that RLHF-based alignment training enhances performance across nearly all NLP evaluation tasks and is fully compatible with domain-specific skills such as Python programming and summarisation. Xiong \textit{et al.}~\cite{xiong2023iterative} reinterpreted RLHF from an information theoretic perspective, proposing an iterative optimization framework that controls alignment bias via KL regularization and drives data acquisition based on uncertainty estimates. SPO~\cite{swamy2024minimaximalist} modeled human preference as a minimax winner in a zero-sum game and directly optimized the policy through self-play, providing theoretical guarantees of robust convergence under non-transitive, non-Markovian, and stochastic preferences. Wang, Fu, and Miao \textit{et al.}~\cite{wang2025beyond, fu2025reward, miao2024inform, miao2025energy} explored methods for mitigating the issue of reward hacking.

Bai \textit{et al.}~\cite{bai2022constitutional} proposed Constitutional AI, a framework in which artificial intelligence systems supervise other AI agents to train a harmless assistant through self-improvement without relying on human-labelled data identifying harmful outputs. The only human oversight is provided through a predefined set of rules or principles. RLAIF~\cite{lee2023rlaif} (Reinforcement Learning with AI Feedback) leveraged existing large language models to generate preference labels without the involvement of human annotators, achieving performance improvements comparable to those of RLHF.

Traditional RLHF requires first training a reward model and then optimizing the policy through RL. This two-stage process is complex and may be unstable. In 2023, Rafailov et al.~\cite{rafailov2023direct} introduced the Direct Preference Optimization (DPO) method, a training paradigm that eliminates the need for explicit reinforcement learning. It has been proven that under certain assumptions, it can bypass explicit reward modeling and RL optimization, and directly fine-tune the pre-trained language model to the optimal policy indicated by preference data through a simple loss function. Azar \textit{et al.} proposed a unified theoretical framework $\Psi$PO~\cite{azar2024general}, which systematically characterizes the fundamental connections and limitations between RLHF and DPO. Smaug~\cite{pal2024smaug} provided a theoretical analysis showing that DPO can fail to preserve the likelihood of preferred sequences when the preference data involves small edit distances. To address this, they introduced a regularized variant, DPOP, which augments the loss with a lower bound constraint on the preferred sequence likelihood. $\beta$-DPO~\cite{NEURIPS2024_ea888178} improved the robustness and alignment performance of DPO under diverse data conditions by dynamically adjusting the KL regularization coefficient $\beta$ based on intra-batch preference quality and employing $\beta$-guided data filtering.

Kwon \textit{et al.}~\cite{kwon2023reward} proposed a method that leverages LLMs as proxy reward functions by generating reinforcement learning signals from natural language prompts, thereby enabling efficient training of agents aligned with user intentions. Eureka~\cite{ma2023eureka} achieved automated reward function design without task-specific prompting by incorporating environment source code as contextual input, combined with reflective reward mechanisms and evolutionary search. Text2Reward~\cite{xie2023text2reward} utilized LLMs to automatically generate interpretable and dense reward functions from natural language instructions. KTO~\cite{ethayarajh2024kto} directly optimized for human utility over generated content rather than the log-likelihood of preferences, achieving performance comparable to or better than existing methods using only binary feedback. ORPO~\cite{hong2024orpo} integrated Supervised Fine-Tuning (SFT) with dynamic preference optimization, significantly enhancing instruction following capabilities and generation quality of language models without requiring a reference model.
Since comprehensive surveys on classical RL-based alignment methods have already been published, this subsection provides only a brief overview of representative approaches during the alignment phase. The latest advances in reward model design will be discussed in detail in the next subsection.

\subsection{Emerging Methods for Reward Model Design}~\label{sec3.3}
The reward model plays a key role in guiding large language models to generate outputs that align with human expectations. Recently, a series of studies~\cite{guo2025reward, zhao2025genprm, chen2025rm, wang2025unified, liu2025inference} have utilized testing-phase computational resources to enhance the performance of reward models. Guo \textit{et al.} present the Reward Reasoning Model (RRM)~\cite{guo2025reward}, which is trained using a reinforcement learning framework. Through the Chain-of-Thought (CoT) reasoning mechanism, RRM can adaptively call additional testing phase computational resources for complex queries where reward judgments are unclear. Zhao \textit{et al.}~\cite{zhao2025genprm} introduced a generative process reward model, which performs explicit CoT reasoning and code verification before making judgments for each reasoning step. Chen \textit{et al.}~\cite{chen2025rm} framed reward modeling as a reasoning task. By distilling high-quality reasoning chains and training in two phases—reinforcement learning based on verifiable rewards—the model achieves state-of-the-art performance on three major reward model benchmarks. Wang \textit{et al.}~\cite{wang2025unified} presented the first unified reward model, UNIFIEDREWARD-THINK, based on multimodal CoT reasoning. This model uses exploration-driven reinforcement fine-tuning to unlock the model's latent complex reasoning capabilities. Liu \textit{et al.}~\cite{liu2025inference} proposed a rule-based online RL trained Pointwise Generative Reward Modeling (GRM) method, which can evaluate single, paired, and multiple responses, thereby overcoming the limitations of traditional reward models in input flexibility. At the same time, Yu \textit{et al.}~\cite{yu2025rewardanything} explored how to improve the generalization of reward models, advocating that reward models should understand and follow dynamically provided natural language reward principles, similar to instruction following in large language models. A novel reward model is introduced that is designed and trained by explicitly following natural language principles. Li \textit{et al.}~\cite{li2025generalist} found that any LLM trained using standard next-token prediction inherently contains a powerful general-purpose reward model. AUTORULE~\cite{wang2025autorule} builds rule-based rewards by extracting rules from preference feedback through interpretation, candidate selection, and merging, then using a verifier to measure rule satisfaction as an auxiliary reward alongside a learned reward model. 
\section{Reinforcement Learning Methods in the Reasoning Phase}~\label{sec4}
\begin{figure}[ht!]
  \centering
  \includegraphics[width=\linewidth, height=\textheight, keepaspectratio]{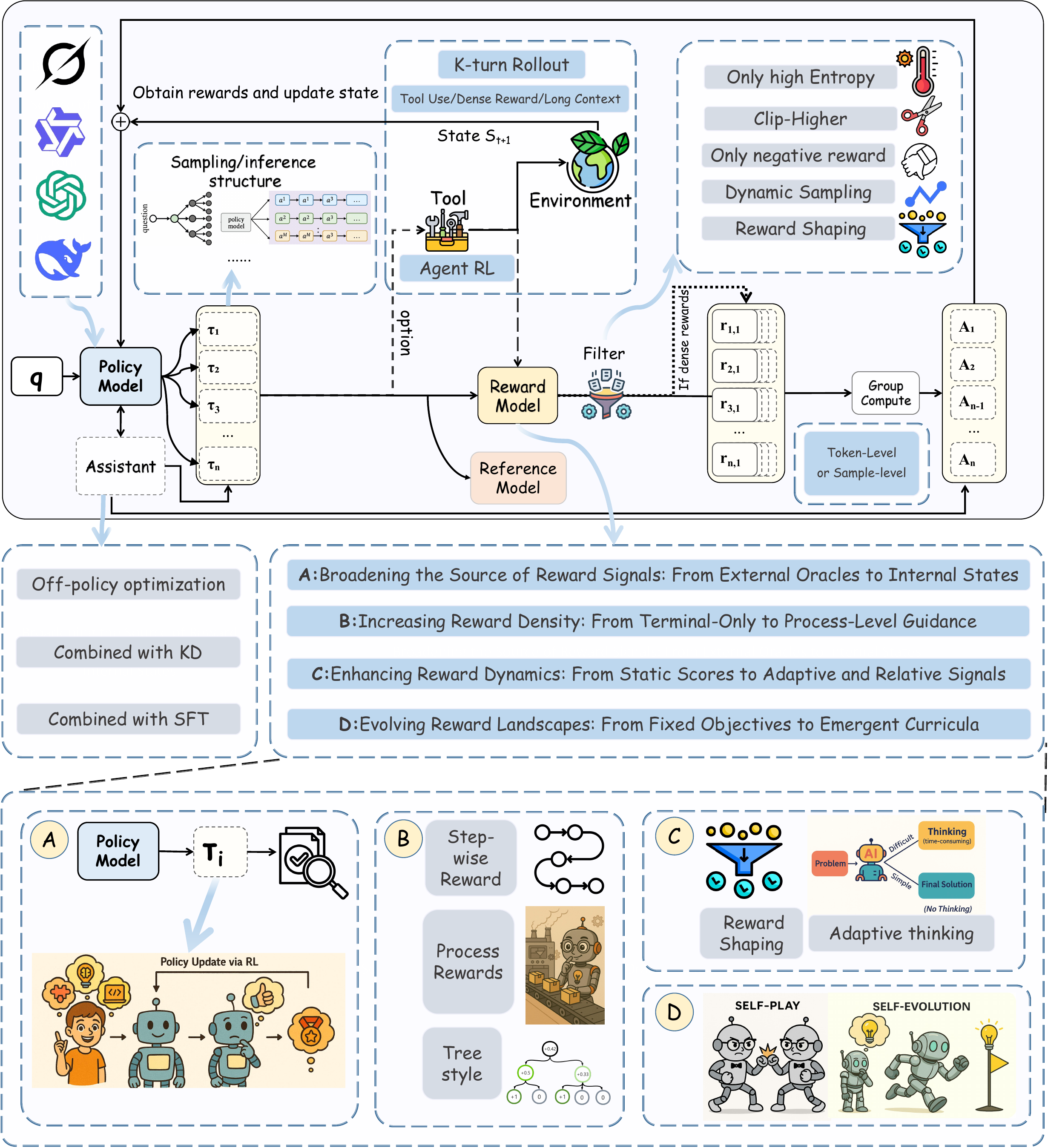}
  \caption{Technical architecture of the RLVR methods. It depicts the overall workflow of the RLVR and expands on the design methods for the reward model, off-policy assistance, reward filtering, sampling and reasoning strategies, Agent RL, and reward update hierarchy.}
  \label{rlvr_arc}
\end{figure}

With the release of GPT-o1~\cite{jaech2024openai} and DeepSeek R1~\cite{guo2025deepseek}, the focus of research on reinforcement learning for large language models has gradually shifted towards RLVR technology in 2025. This chapter will introduce the latest advancements in the algorithms of RLVR technology and will discuss its applications in multimodal reasoning, adaptive thinking, agents, as well as the integration of RL with other techniques in the fine-tuning phase. Figure \ref{rlvr_arc} illustrates the overall framework diagram of RLVR technology and the key technical points that can be improved.
\subsection{Experimental Findings of RLVR in Improving the Reasoning Ability of LLMs}~\label{sec4.1}
Reinforcement Learning with Verifiable Rewards has recently demonstrated notable success in enhancing the reasoning performance of LLMs, particularly in mathematics and programming tasks. However, there is controversy in academic circles over whether RL truly expands the model's reasoning ability, or merely amplifies the high-reward outputs already present in the base model's distribution, as well as whether continuously increasing computational power for reinforcement learning can reliably improve reasoning performance. Liu \textit{et al.}~\cite{liu2025prorl} found that when sufficient training time is provided and applied to new reasoning tasks, reinforcement learning can indeed discover entirely new solution paths that the base model lacks entirely. Although RLVR improves the sampling efficiency of correct reasoning paths, Yue \textit{et al.}~\cite{yue2025does}, using pass@k (the probability of generating at least one correct solution among k samples) as an evaluation metric, revealed that the current training paradigm has not stimulated truly novel reasoning patterns. Data shows that RLVR models outperform the base model when k is small (\textit{e.g.}, k=1), but the base model performs better when k increases. The boundary of reasoning ability in LLMs often narrows as RLVR training progresses. Analysis of coverage and perplexity indicates that all reasoning paths generated by RLVR exist in the sampling distribution of the base model, suggesting that its capabilities are inherently derived from and limited by the base model. Prior to this, several studies ~\cite{liu2025there, zhao2025echo, shah2025rethinking} have pointed out that the reflective behavior in RLVR models stems from the base model itself, rather than being acquired through reinforcement learning. To address the above controversies, Wu \textit{et al.}~\cite{wu2025invisible} pointed out that RLVR is mainly an efficient sampler. Although it can occasionally explore capabilities beyond the base model, it often fails to solve new problems due to diversity collapse, and it also forgets the problems that the base model already knows how to solve. Regarding the phenomenon of policy entropy collapse that continuously occurs in massive reinforcement learning experiments without entropy intervention, Cui \textit{et al.}~\cite{cui2025entropy} established a conversion equation between entropy value $H$ and downstream performance $R$:
\begin{equation}
    R=-a\exp(\mathcal{H})+b.
\end{equation}
This equation indicates that policy performance is achieved at the cost of entropy consumption, and thus is limited by the depletion of entropy, with a fully predictable theoretical upper limit ($R=-a+b$ when $H=0$). This finding suggests that continuous exploration must be achieved through entropy management to break through the computational power expansion bottleneck of reinforcement learning.

There is disagreement regarding whether to retain entropy regularization~\cite{ouyang2022training, shao2024deepseekmath, hu2025open}. Through experiments, Cui \textit{et al.}~\cite{cui2025entropy} showed that although entropy values need to be controlled, it is possible to design objective functions that are superior to entropy loss. The study found that the covariance exhibited by a small number of tokens is extremely high, far exceeding the average level. This means that these abnormal tokens play a dominant role in triggering entropy collapse. Separating the gradients of high-covariance tokens and preventing their updates from reducing entropy can suppress entropy collapse. Coincidentally, Wang \textit{et al.}~\cite{wang2025beyond} found that only a few high-entropy tokens guide the model's reasoning paths, and training on just 80\% of low-entropy tokens significantly reduces performance. High-entropy tokens (\textit{e.g.}, ``however,'' ``thus,'' ``because,'' ``suppose'') handle logical connections, while low-entropy tokens (\textit{e.g.}, affixes, code snippets, mathematical expression components) are used for sentence construction, with other tokens blending both roles. Limiting policy gradient updates to branching tokens can improve RLVR.

Li \textit{et al.}~\cite{li2025think} systematically investigated the necessity of explicit thinking processes in rule-based reinforcement fine-tuning for multimodal large language models. Research has shown that for some tasks and models, removing or adjusting the thinking process can actually improve performance and efficiency. Ma \textit{et al.}~\cite{ma2025reasoning} questioned the necessity of explicit thinking. This study reveals that using simple prompts to bypass the thinking process yields better results, and it is found that as the k-value increases, the competitiveness of the non-thinking method in the pass@k metric continues to strengthen. Samineni \textit{et al.}~\cite{samineni2025rl} critically analyzed the structural assumptions of reinforcement learning in the post-training of large language models, pointing out that these assumptions reduce RL to filtered iterative supervised fine-tuning. Through theoretical and empirical evidence, it reveals that the increase in response length is a side effect of training settings rather than an improvement in reasoning ability. Zhu \textit{et al.}~\cite{zhu2025surprising} decomposed learning signals into two categories: reinforcing correct answers and punishing incorrect ones. Subsequently, it is found that training using only negative samples without reinforcing correct answers may be more effective. Bogdan \textit{et al.}~\cite{bogdan2025thought} proposed three methods—black box resampling, attention aggregation for receiver heads, and causal suppression—to identify ``thought anchors'' pivotal sentences guiding LLM CoT reasoning, validated by the case study and visualization tool. Wu \textit{et al.}~\cite{wu2025reasoning} systematically analyzed that Qwen2.5~\cite{qwen2025qwen25technicalreport} conducted RLVR training on the model by constructing a clean mathematical dataset under different reward settings. The authors found that Qwen can truly improve its mathematical reasoning ability only when trained on clean data with accurate rewards, emphasizing the importance of data cleanliness and the quality of reward design in evaluating RL methods. Chen \textit{et al.}~\cite{chen2025sft} analyzed the limitations of the traditional ``supervised fine-tuning + reinforcement learning'' strategy and found that SFT is prone to generating ``pseudo reasoning paths''. This not only fails to effectively enhance complex reasoning abilities but also significantly impairs the performance of the subsequent RL stage. He \textit{et al.}~\cite{he2025response} presented the Trajectory Policy Gradient Theorem (TPGT), showing response-level rewards can estimate token-level rewards for LLM reinforcement learning, introducing the efficient TRePO algorithm, and comparing it to methods like PPO and DPO. Shojaee \textit{et al.}~\cite{shojaee2025illusion} investigated Large Reasoning Models through controlled puzzle environments, demonstrating their effectiveness in moderately complex tasks but revealing significant limitations, including accuracy collapse at high complexity and inefficient reasoning processes, questioning their generalizable reasoning capabilities. Liu \textit{et al.}~\cite{liu2025more} found that stronger reasoning often increases hallucinations, as longer chains shift attention from image content to language priors, with attention analysis showing weakened visual focus exacerbates this effect.

\subsection{Recent Advances in RL Algorithms for LLMs}~\label{sec4.2}
This section chronologically introduces the main algorithms of RLVR, focusing on Group Relative Policy Optimization (GRPO) ~\cite{shao2024deepseekmath} and its improved algorithms. As RL training for large models on long-chain reasoning tasks such as mathematics competitions and code generation has matured, new challenges and corresponding algorithms have emerged. ByteDance and collaborators released Decoupled Clip and Dynamic Sampling Policy Optimization (DAPO)~\cite{yu2025dapo}, an open-source framework for large-scale long-sequence RL training of LLMs. DAPO, built on GRPO, enhances long-CoT performance with four techniques: Clip-Higher, Dynamic Sampling, Token-Level Policy Gradient Loss, and Overlong Reward Shaping. Unlike PPO/GRPO’s symmetric clipping that restricts low probability actions and causes entropy collapse, DAPO relaxes the upper bound to maintain exploration. It resamples extreme reward prompts to avoid wasted samples and speed up convergence. By weighting loss by sequence length, it discourages verbose, low-quality outputs while retaining high-quality, long ones. Finally, it penalizes or truncates overly long outputs, curbing uncontrolled growth and stabilizing training.

Open-Reasoner-Zero~\cite{hu2025open} adopted a minimalist training strategy to achieve efficient and scalable improvement in reasoning ability on foundation models without pre-training fine-tuning, significantly outperforming DeepSeek-R1-Zero with only one-tenth of the training steps. Zhang \textit{et al.}~\cite{zhang2025distill} generated pseudo-rewards in a self-supervised manner by leveraging the intrinsic structure of responses from teacher and student models, enabling reward learning without explicit external evaluation. Kimina-Prover~\cite{wang2025kimina} is an RL-trained LLM that enhances reasoning capabilities in Lean 4 theorem proving by constructing structured formal reasoning patterns, achieving performance improvements that scale with model size without relying on external search algorithms. SRPO~\cite{zhang2025srpo} enhanced the reasoning capabilities of LLMs in mathematics and programming tasks through a two-stage reinforcement learning–centric training strategy combined with a historical resampling mechanism, providing a viable pathway for improving cross-task reasoning capabilities. Yan \textit{et al.}~\cite{yan2025learning} introduced off-policy reasoning trajectories from external models and, through a combination of mixed-policy optimization and regularized importance sampling, enabled the model to learn from both its own generated data and external demonstrations during training. X-Reasoner~\cite{liu2025x} through general text post-training (\textit{i.e.,} SFT + RL only), has been experimentally proven to have reasoning abilities that can generalize across modalities and domains. TANGO~\cite{zha2025rl} jointly trained the generator and generative process-level verifier of large language models through reinforcement learning, aiming to enable the two to promote each other and evolve synergistically without step-by-step annotations. Zhou \textit{et al.}~\cite{zhou2025reinforcing} extended R1-Zero-style training to tasks without rule-verifiable answers by generating reasoning trajectories, concatenating them with reference answers, and evaluating the likelihood of the reference answer. This likelihood serves both as a reward for trajectory optimization and as a weight for supervised training. REINFORCE++~\cite{hu2025reinforce++} achieves PPO-like training stability and efficiency without relying on a value network, by incorporating clipped policy updates, KL divergence penalties, and advantage normalization.

SuperRL~\cite{liu2025superrl} detected reward sparsity via an adaptive switching mechanism and, when sparsity is identified, activates a hybrid executor that combines policy gradients with offline supervision to stabilize learning.
KDRL~\cite{xu2025kdrl} explored the integration of teacher supervision and RL by constructing a unified objective function that incorporates GRPO and KD. Graph-R1~\cite{luo2025graph} proposed a proxy-based GraphRAG framework that enhances the accuracy, efficiency, and generation quality of LLMs in knowledge-intensive tasks through lightweight knowledge hypergraph construction, multi-turn interactive retrieval, and end-to-end reinforcement learning optimization. R2-Reasoner~\cite{shao2025route} employed a reinforced router to decompose queries and allocate subtasks across heterogeneous LLMs, enabling collaborative reasoning that balances accuracy, efficiency, and cost. ToTRL~\cite{wu2025totrl} improved the reasoning efficiency of language models in multi-path logical problems by guiding them to transition from linear chain reasoning to tree-structured thinking. TreeRPO~\cite{yang2025treerpo} constructs tree-structured reasoning paths and optimizes relative node rewards via tree-based sampling, providing LLMs with step-level dense feedback without a reward model. TreeRL~\cite{hou2025treerl} introduced the entropy-based tree search strategy EPTree into the reinforcement learning training process of large language models to improve the diversity of reasoning paths and the quality of process supervision signals. 

Magistral~\cite{rastogi2025magistral} proposed a method for enhancing the reasoning capabilities of large language models through reinforcement learning alone, without distilling reasoning trajectories. MiniMax-M1~\cite{chen2025minimax} introduced a large language model with a mixture-of-experts architecture and a Lightning Attention mechanism, incorporating the CISPO reinforcement learning algorithm to improve training stability and efficiency in RL, while supporting million-token contexts and enabling efficient inference. SPIRAL~\cite{liu2025spiral} introduced a multi-agent self-play RL framework with zero-sum language games that improves LLM reasoning without supervision or domain data by generating evolving tasks through adversarial opponents and stabilizing training via role-conditioned advantage estimation. The GSPO algorithm~\cite{zheng2025group} improved RL training stability and efficiency by replacing token-level importance weights with sequence-level ratios for reward computation and policy updates, alleviating granularity mismatch.
\subsection{Application of RLVR in Multimodal Reasoning}~\label{sec4.3}
DeepSeek-R1-Zero has successfully demonstrated that reasoning capabilities can emerge in large language models solely through reinforcement learning. A large body of work has begun to explore how to leverage reinforcement learning to foster multimodal reasoning abilities. Vision-R1~\cite{huang2025vision} took the lead in applying reinforcement learning to enhance the reasoning capabilities of Multimodal Large Language Models (MLLMs), while systematically analyzing the differences between direct reinforcement learning training and the combined approach of ``cold-start initialization + reinforcement learning training''. Visual-RFT~\cite{liu2025visual} proposed a reinforcement fine-tuning framework for Vision-Language Models (VLMs), which is applied in fields such as detection, localization, and classification. VLM-R1~\cite{shen2025vlm} observed that many vision understanding tasks have explicitly annotated ground truths, making them suitable for rule-based reward mechanisms, and thereby extended R1-style reinforcement learning to vision–language models. R1-onevision~\cite{yang2025r1} designed a cross-modal reasoning process, which converts images into standardized text representations, thereby enabling precise language-based reasoning. Deng \textit{et al.}~\cite{deng2025boosting} specifically designed a curriculum reinforcement fine-tuning algorithm for small-scale Vision-Language Models. Ma \textit{et al.}~\cite{ma2025deepperception} explored integrating reasoning capabilities into visual perception. Zhou \textit{et al.}~\cite{zhou2025r1} successfully reproduced the emergent properties of multimodal reasoning for the first time on an unsupervised fine-tuned model with only 2 billion parameters. Through experiments, VisuLogic~\cite{xu2025visulogic} was found that RL is an effective way to enhance the visual reasoning ability of multimodal large language models. R1-VL~\cite{zhang2025r1} proposed the StepGRPO algorithm, which provides rewards for paths containing necessary intermediate reasoning steps through soft key step matching technology, and incentivizes reasoning processes that are structured and logically consistent through reasoning completeness and logicality evaluation strategies, thereby achieving dense rewards for visual-language reasoning tasks. SophiaVL-R1~\cite{fan2025sophiavl} incorporated reasoning-process reward signals by computing the difference in reasoning rewards between correct and incorrect answers, and dynamically assigning confidence weights to these reasoning rewards, thereby mitigating the impact of unreliable reasoning rewards.

Video-R1 ~\cite{feng2025video} and TinyLLaVA-Video-R1 ~\cite{zhang2025tinyllava} successively applied reinforcement learning to the video domain. Video-R1 established a dataset for video reasoning and designed a mechanism where the model receives positive rewards only when its current reasoning strategy for specific questions demonstrates dependence on temporal information, thereby strengthening the large model's temporal modeling ability. TinyLLaVA-Video-R1~\cite{zhang2025tinyllava}, on the other hand, explored how to use reinforcement learning to enhance the video reasoning capabilities of small-scale models. Liao \textit{et al.}~\cite{liao2025improved} discussed methods to enhance the visual-spatial reasoning abilities of MLLMs through R1-Zero-style training. SpaceR~\cite{ouyang2025spacer} extended GRPO through a map imagination mechanism, prompting the model to infer spatial layouts during the thinking process, thereby enhancing the video spatial reasoning ability of MLLMs. The essence of humans' and robots' cognition of the environment lies in perceiving and understanding spatial relationships through first-person perspective video streams. Therefore, a crucial aspect of embodied intelligence tasks is that the model needs to possess the ability to perceive and understand spatial relationships from first-person perspective video streams. Zhao \textit{et al.}~\cite{zhao2025embodied} proposed the Embodied-R framework, which realizes collaborative work by combining the perceptual capabilities of large-scale VLMs with the reasoning capabilities of small-scale Language Models. Ego-R1 ~\cite{tian2025ego} explored a new framework for reasoning over ultra-long (measured in days/weeks) first-person videos. By decomposing complex reasoning into modular steps, the RL agent iteratively collaborates by calling specific tools at each step, sequentially solving sub-problems such as temporal retrieval and multimodal understanding, thereby significantly extending the time coverage from several hours to a week. VAU-R1 ~\cite{zhu2025vau} extended RLVR to the field of Video Anomaly Understanding (VAU), enhances anomaly reasoning capabilities through Reinforcement Fine-Tuning (RFT), and introduces VAUBench, the first CoT benchmark specifically designed for video anomaly reasoning.  VLN-R1~\cite{qi2025vln} extended RLVR to the field of Vision-Language Navigation (VLN). CAD-Coder~\cite{guan2025cad} introduced RL into the field of artificial intelligence-assisted CAD. Chen \textit{et al.}~\cite{chen2025compile} combined SFT with GRPO and introduce a rule-based reward tailored to scene graph structures, improving the structural validity, relationship recall, and long-tail category performance of MLLMs in end-to-end scene graph generation. ARMed~\cite{liu2025breaking} explored the application of RLVR techniques in the domain of medical imaging. 3D-R1~\cite{huang20253d} enhanced the reasoning and generalization capabilities of 3D VLM by combining cold-start initialization with high-quality CoT datasets, RL–based training, and a dynamic view selection strategy.

Huang \textit{et al.}~\cite{huang2025boosting} proposed Hint-GRPO to address GRPO’s low data utilization—where models struggle to gain positive updates from hard samples—and text bias—where models ignore images and rely only on text—by supplying partial correct reasoning steps as hints and leveraging prediction differences between inputs with and without image conditions to enforce visual grounding. SRPO ~\cite{wan2025srpo} explored the problem that MLLMs still lag significantly behind unimodal text models in complex problems that require explicit self-reflection and correction. Wang \textit{et al.}~\cite{wang2025vicrit} extended the successful pattern of RL in mathematical reasoning and code generation to the field of visual perception by injecting subtle synthetic visual hallucinations into manually written image description paragraphs and training VLMs to locate these errors.

T2I-R1 ~\cite{jiang2025t2i} took the lead in introducing RLVR strategies into the field of visual generation, proposing a new text-to-image generation model based on a two-level CoT reasoning framework and reinforcement learning, which even surpassed FLUX.1 on benchmarks. DanceGRPO ~\cite{xue2025dancegrpo} is the first unified framework that adapts GRPO to visual generation paradigms. It enables the universal deployment of a single reinforcement learning algorithm across diffusion models and rectified flows generative paradigms; text-to-image, text-to-video, and image-to-video tasks; foundation models such as Stable Diffusion, HunyuanVideo, FLUX, and SkyReels-I2V; and reward models including image/video aesthetics, text-image alignment, video motion quality, and binary reward. Xu \textit{et al.}~\cite{xu2025visual} explored planning through visual representations completely independent of text, confirming that pure visual planning can serve as a feasible alternative to language-based reasoning.

Although RL has already been applied to multimodal reasoning, its reasoning process is still mainly limited to text forms~\cite{yang2025r1}. There is still substantial room for exploration in approaches that deeply integrate multimodal information and external tools. GPT o3 ~\cite{o3-o4-mini-system-card} has demonstrated a strong ability to reason by relying on visual cues, setting off a new upsurge in visual reasoning research, with researchers scrambling to explore methods to achieve GPT o3-style visual reasoning. DeepEyes ~\cite{zheng2025deepeyes} proposed an interleaved multimodal reasoning paradigm, where the model decides at each step whether to continue reasoning with text or call tools to crop regions of the image as historical context, and proceeds with reasoning in this way to form an interleaved reasoning sequence. Several methods have been proposed to address the limitation that models often struggle to effectively anchor their reasoning on visual cues. For instance, GThinker~\cite{zhan2025gthinker} introduced CueRethinking, which anchors the inference process to visual cues and resolves inconsistencies through iterative cue reinterpretation. GLM-4.1V-Thinking~\cite{hong2025glm} proposed a unified multimodal reasoning framework that integrates multi-source pre-training, supervised fine-tuning, and cross-task reinforcement learning with RLCS and a refined multi-domain reward system to enhance reasoning across diverse tasks.
\subsection{Adaptive Reasoning}~\label{sec4.4}
RLVR technology has achieved improved performance during testing by consuming more computing resources to generate longer chain-of-thought sequences. However, the length of its CoT reasoning is uncontrollable, making it impossible to achieve the expected performance level by allocating computing resources during testing. It often occurs that excessive testing time is allocated to simple problems or insufficient testing time is allocated to difficult problems~\cite{aggarwal2025l1, yi2025shorterbetter, lou2025adacot}. This has inspired researchers to study adaptive length reasoning methods for large language models. S1 ~\cite{muennighoff2025s1}extended computation through ``budget forcing'' technique. L1 ~\cite{aggarwal2025l1} proposed Length Controlled Policy Optimization, an RL-based method. After training, the model can generate outputs that meet the required length constraints given in the prompts, and its performance is better than that of S1. Yi \textit{et al.}~\cite{yi2025shorterbetter} defined Sample Optimal Length as the length of the shortest correct response among multiple generated results, and uses it as a dynamic reward signal to guide the model to achieve efficient reasoning. AdaCoT ~\cite{lou2025adacot}, based on the PPO method, dynamically controlled the CoT trigger decision boundary by adjusting the penalty coefficient, enabling the model to judge the necessity of CoT according to the implicit query complexity. Thinkless ~\cite{fang2025thinkless} is trained under the reinforcement learning paradigm and uses two control symbols: <short> (concise) and <think> (deliberate). It decomposes the hybrid reasoning learning objective into a control symbol loss function (managing the selection of reasoning modes) and an answer loss function, thereby enhancing the hybrid-length reasoning ability. AdaptThink~\cite{zhang2025adaptthink} showed that skipping reasoning benefits simple tasks in performance and efficiency, and by constraining the objective with importance sampling, it balances “thinking” and “non-thinking” samples during training to enable adaptive mode selection. Jiang \textit{et al.}~\cite{jiang2025think} designed a two-stage training framework of first cold start and then reinforcement learning to realize a model that can adaptively judge whether to start the thinking process according to the context information of user queries. Zhang \textit{et al.}~\cite{zhang2025continue} systematically quantified the performance upper bounds of Large Reasoning Models (LRMs) in ``long thinking'' and ``non-thinking'' modes. By introducing a precision-aware length reward adjustment mechanism, it adaptively allocates reasoning resources according to problem difficulty to achieve efficient reasoning. Wang \textit{et al.}~\cite{wang2025adaptive} first performed supervised fine-tuning on the base model to simultaneously acquire long and short chain reasoning abilities. Then, it adopted a long-short adaptive grouping reward strategy to evaluate the prompt complexity and give corresponding incentives, and implements a logic-based reasoning mode switching loss function to optimize the initial token selection of the model, thereby guiding the reasoning type decision. Zhang \textit{et al.}~\cite{zhang2025alphaone} used the parameter $\alpha$ to characterize the scaled thinking phase. After the $\alpha$ moment ends, $\alpha1$ deterministically terminates slow thinking through a thinking termination marker, thereby promoting rapid reasoning and efficient answer generation. This study models the insertion of reasoning transition markers as a Bernoulli random process to dynamically schedule slow-thinking transitions.
\subsection{RLVR for Agent}~\label{sec4.5}
The interplay between long-horizon decision-making and stochastic environmental feedback introduces unique challenges in training large language models as interactive agents. Unlike typical single-turn interaction tasks common in standard LLM applications, interactive agents necessitate extended, multi-turn interactions with their environment. Despite the advancements of reinforcement learning in static tasks, multi-turn agent training in reinforcement learning remains understudied, particularly due to issues such as delayed rewards ~\cite{li2024larm, feng2025group, wang2025spa}. AGILE ~\cite{peiyuan2024agile} integrated large language models with memory systems, tool utilization, and expert interactions. Here, the LLM serves as the policy model and is fine-tuned using annotated action data through the Proximal Policy Optimization algorithm. LARM ~\cite{li2024larm} employed lightweight LLMs (fewer than 5 billion parameters) to directly output executable actions rather than textual descriptions. Utilizing the PPO algorithm for training and introducing a referee mechanism based on large-scale LLMs for providing intermediate rewards, LARM successfully addresses the challenge of long-term reward vanishing, demonstrated by obtaining enchanted diamond equipment in Minecraft. Search-R1~\cite{jin2025search} optimized the reasoning trajectory of LLMs through multi-round search interactions and adopts retrieval token masking technology to ensure the stability of reinforcement learning training. ToRL~\cite{li2025torl} extended reinforcement learning directly from foundational models (\textit{i.e.}, without additional post-training), enabling large language models to autonomously leverage computational tools. ReTool~\cite{feng2025retool} enhanced long-chain reasoning by generating synthetic cold-start data to build code-enhanced reasoning trajectories for fine-tuning, and then applying iterative reinforcement learning with task success rewards to autonomously optimize tool-use strategies without manual priors. RAGEN~\cite{wang2025ragen} investigated four environments and identified the “Echo Trap” of reward fluctuations and gradient spikes, highlighting that robust reasoning in multi-turn RL requires diversified initial states, moderate interaction granularity, and higher sampling frequency, while lacking fine-grained reasoning-aware rewards leads agents to superficial or hallucinatory strategies. OpenThinkImg ~\cite{su2025openthinkimg} optimized task success rates directly through interactive tool feedback, enabling large vision-language models (LVLMs) to autonomously identify optimal tool-use strategies. Feng \textit{et al.}~\cite{feng2025group} addressed the sparse and delayed reward issues emerging from extended multi-step interactions by computing macro-level relative advantages at the episodic level based on full trajectory groups, and step-level groupings through an anchored state grouping mechanism. Zhang \textit{et al.}~\cite{zhang2025divide} adopted a heuristic approach to group games based on features such as rules and difficulty, subsequently training specialized models for each group. It then merges the parameters from these group-specific models into a unified model, which is further trained across multiple groups until effective generalization across diverse game scenarios is achieved. Tool-Star ~\cite{dong2025tool} proposed a two-stage training framework integrating six tool categories, employing a hierarchical reward design through a multi-tool self-assessment RL algorithm. SPA-RL~\cite{wang2025spa} decomposed final rewards into stepwise contributions aligned with task progress, used a progress evaluator to match cumulative values to completion, and combined them with baseline signals to yield fine-grained intermediate rewards that mitigate delay and enhance training. Shop-R1~\cite{zhang2025shopr1rewardingllmssimulate} improved LLMs’ shopping behavior simulation by rewarding action correctness proportional to difficulty, while noting that limited context windows constrain long-term dynamic reasoning. Memory-R1~\cite{yan2025memory} learned to manage external memory and utilize it for long-term reasoning through two dedicated agents: a memory manager and a response agent. MemAgent~\cite{yu2025memagent} employed segmented processing and selective memory mechanisms to manage extended contextual information. In contrast, RMM~\cite{tan2025prospect} and M3-Agent~\cite{long2025seeing} explored strategies for managing multi-turn long-term memory.

\subsection{Reinforcement Learning from Internal Feedback}~\label{sec4.6}
While methods such as RLHF and RLVR have achieved remarkable results, they require extensive external supervision. However, a series of studies~\cite{yue2025does, wu2025invisible} have found that RLVR does not inspire truly novel reasoning patterns; it merely improves the sampling efficiency of correct reasoning paths. Since RLVR does not bring new external information to LLMs but only stimulates the knowledge learned during pre-training, can we find a way to enable LLMs to activate such pre-trained knowledge without external supervision? Kang \textit{et al.}~\cite{kang2025scalable} hypothesized that higher distributed self-certainty across samples correlates with response accuracy and thus proposed self-certainty as a metric for evaluating response quality without external rewards. TTRL ~\cite{zuo2025ttrl} utilized prior knowledge from pre-trained models during reinforcement learning training and employed a majority voting method to address the lack of explicitly labeled data, achieving performance improvements. Zhao \textit{et al.}~\cite{zhao2025absolute} proposed a self-evolving model that autonomously generates and solves tasks to maximize learning progress without external data, using a code executor to verify both tasks and solutions and provide unified rewards for open-ended learning. SLOT~\cite{hu2025slot} is a lightweight test-time method that, without altering the main model, optimizes an additive vector $\delta$ for each prompt using self-constructed supervision signals to minimize language modeling loss and thereby improve accuracy and reasoning under complex instructions. Zhao \textit{et al.}~\cite{zhao2025learning} used the model's own confidence as the sole reward signal and replaced the external reward in GRPO with a self-certainty score, achieving fully unsupervised learning. Kang \textit{et al.}~\cite{kang2025scalable} hypothesized that higher aggregated self-certainty across samples correlates with response accuracy and proposed it as a metric for evaluating quality without external rewards. Li \textit{et al.}~\cite{li2025confidence} used the model's own confidence as the reward signal, eliminating the need for manual annotation, preference models, or reward function design. Zhang \textit{et al.}~\cite{zhang2025no} showed that RLIF, using unsupervised reward proxies (entropy and self-certainty), initially boosts base LLM reasoning to rival RLVR but later degrades below pre-finetuning levels, yields limited gains for instruction-tuned models, and exposes intrinsic causes of these behaviors. RLSF~\cite{vanniekerk2025posttraininglargelanguagemodels} builds preference data from models’ self-evaluated answer confidence and fine-tunes them with RL to improve calibration and reasoning in math and multiple-choice tasks.
\section{Datasets and Benchmarks}~\label{sec5}
\subsection{Synthetic Data Generation}~\label{sec5.1}
Zhu \textit{et al.}~\cite{zhu2025data} proposed a synthetic data framework for abstract visual reasoning that generates structured QA pairs and reasoning chains via A-SIG for regular patterns and crawls, templates, and manual annotations for irregular ones, providing diverse, well-defined samples to enhance both perception and reasoning. Goldie \textit{et al.}~\cite{goldie2025synthetic} designed a synthetic data pipeline for multi-step reasoning and tool use by leveraging LLM–tool interactions to build stepwise trajectories with context, actions, and feedback, and applying filtering based on process plausibility and final correctness to yield high-quality offline data for reinforcement learning. Guo \textit{et al.}~\cite{guo2025synthetic} proposed a task-definition-based synthetic data RL method that generates QA pairs from task definitions, adapts difficulty to model performance, and reinforces with high-potential samples solvable but not yet mastered, enabling task adaptation without human labels. SwS~\cite{liang2025sws} targeted reasoning tasks in reinforcement learning with large language models. It identifies problems that the model consistently fails to solve during training, extracts the underlying concepts involved, and synthesizes targeted follow-up questions to enhance training. 

\subsection{Datasets and Benchmarks}~\label{sec5.2}
This section will introduce the common datasets and test benchmarks in reinforcement learning for large language models. Table \ref{tab:benchmark_overview} summarizes commonly used datasets and benchmarks spanning alignment/dialogue, code, math, and general knowledge.
\begin{table*}[t]
  \centering
  \caption{Datasets and Benchmarks Overview. This table categorizes and lists common datasets and benchmarks for RL-LLM research, covering fields such as alignment, code, mathematics, knowledge, logical reasoning, and agentic tasks.}
  \label{tab:benchmark_overview}
  \setlength{\tabcolsep}{6pt}     
  \renewcommand{\arraystretch}{1.2} 
  \footnotesize                   
  \begin{tabularx}{\textwidth}{@{} C{.22\textwidth} C{.38\textwidth} Y @{}}
    \toprule
    \textbf{Category} & \textbf{Dataset \& Benchmark} & \textbf{Summary} \\
    \midrule
    Alignment / Dialogue &
      HHH~\cite{askell2021general}, HH-RLHF~\cite{bai2022training}, IFEval~\cite{zhou2023instruction}, Arena-Hard~\cite{li2024crowdsourced}, AlignBench~\cite{liu-etal-2024-alignbench}, Creative Writing~\cite{creative-writing-bench-v3} &
      Evaluates alignment in dialogue, focusing on helpfulness, honesty, and harmlessness. \\
\midrule
    Code &
      APPS~\cite{dou2024stepcoder}, LiveCodeBench~\cite{jain2024livecodebench}, SWE-bench~\cite{lewkowycz2022solving}, SWE-bench Verified~\cite{yang2025swe}, OJBench~\cite{wang2025ojbench} &
      Programming tasks involve code generation and debugging, with automated or real-time evaluation. \\
\midrule
    Math &
      GSM8K~\cite{cobbe2021training}, MATH~\cite{hendrycks2021measuring}, OlympiadBench~\cite{he2024olympiadbench}, Minerva Math~\cite{lewkowycz2022solving}, OlympiadBench~\cite{he2024olympiadbench}, PolyMath~\cite{wang2025polymath}, AMC2023, AIME2024/2025, CNMO2024, HMMT2025 &
      Benchmarks for solving mathematical problems, from elementary to advanced levels, including competition and Olympiad-level tasks. \\
\midrule
    General Exams / Knowledge \& STEM &
      MMLU~\cite{hendrycks2020measuring}, MMLU-Pro~\cite{wang2024mmlu}, GPQA~\cite{rein2024gpqa}, SuperGPQA~\cite{du2025supergpqa}, TheoremQA~\cite{chen2023theoremqa}, Guru~\cite{cheng2025revisiting}, SimpleQA~\cite{wei2024measuring}, HLE~\cite{phan2025humanity}, LiveBench~\cite{White2025LiveBench}, PhyX~\cite{shen2025phyx}, BBH~\cite{srivastava2023beyond}, BBEH~\cite{kazemi2025big}, MMReason~\cite{yao2025mmreason}&
      General knowledge benchmarks covering various fields, including STEM and human-level exam comparisons. \\
\midrule
    Logic Reasoning &
      AutoLogi~\cite{zhu2025autologi}, ZebraLogic~\cite{lin2025zebralogic} &
      Logic Reasoning Evaluation. \\
\midrule
    Tools / Multi-turn / Agent &
      $\tau^2$-Bench~\cite{barres2025tau}, ACEBench~\cite{chen2025acebench}, MultiChallenge~\cite{sirdeshmukh2025multichallenge} &
      Benchmarks testing multi-turn interaction with tools, agent-based reasoning. \\
    \bottomrule
  \end{tabularx}
\end{table*}
HHH~\cite{askell2021general} is a dialogue-based benchmark designed to evaluate large language models along three critical dimensions: Helpful, Honest, and Harmless (HHH). It is intended to assess the degree of alignment between the model and human expectations during interaction. HH-RLHF~\cite{bai2022training} focused on training preference (or reward) models as a precursor to RLHF. The test benchmarks for alignment tasks also include IFEval~\cite{zhou2023instruction}, Arena-Hard~\cite{li2024crowdsourced}, AlignBench~\cite{liu-etal-2024-alignbench}, Creative Writing~\cite{creative-writing-bench-v3}, and so on. APPS~\cite{hendrycks2021measuring} evaluated a model's ability to generate satisfactory Python code given arbitrary natural language specifications. APPS+~\cite{dou2024stepcoder} built upon the original APPS dataset with manual verification and refinement, containing 7,456 examples, each including a programming task description, reference solution, function signature, unit tests (input/output), and starter code, specifically designed for code generation. LiveCodeBench~\cite{jain2024livecodebench} extended evaluation beyond code generation to broader code-related skills, including self-repair, code execution, and output prediction. GSM8K~\cite{cobbe2021training} contained 8,500 high-quality, linguistically diverse elementary school math word problems. It also demonstrates that verification strategies significantly improve model performance on GSM8K, especially when scaling data. MATH~\cite{hendrycks2021measuring} introduced 12,500 challenging competition-style math problems, each accompanied by detailed step-by-step solutions for training models to produce answer derivations and explanations. MMLU~\cite{hendrycks2020measuring} measured multitask accuracy of language models across 57 subjects, including elementary math, U.S. history, computer science, and law. Achieving high accuracy requires extensive world knowledge and problem-solving abilities. MMLU-Redux~\cite{gema2024we} adopted a novel error annotation protocol to identify errors in the dataset, thereby improving MMLU~\cite{hendrycks2020measuring}. MMLU-Pro~\cite{wang2024mmlu} extended MMLU by incorporating more challenging, reasoning-focused questions and increasing the number of answer choices from four to ten. It also removed ambiguous or noisy items present in the original MMLU. GPQA~\cite{rein2024gpqa} consisted of 448 multiple-choice questions written by domain experts across biology, physics, and chemistry. Expert accuracy is around 65\%, while non-experts (with web access) achieve 34\%, and GPT-4 scores 39\%. Designed to be “Google-proof,” the benchmark evaluated deep scientific reasoning. SuperGPQA~\cite{du2025supergpqa} assessed postgraduate-level reasoning and domain knowledge across 285 disciplines. It employed a novel human–LLM co-filtering strategy to iteratively refine questions using model outputs and expert feedback, eliminating vague or ill-formed items. TheoremQA~\cite{chen2023theoremqa} is the first theorem-driven QA dataset, targeting a model's ability to apply formal theorems to solve complex scientific problems. It includes 800 expert-curated high-quality questions covering 350 theorems from mathematics, physics, electrical engineering, computer science, and finance. Guru~\cite{cheng2025revisiting} is a reinforcement learning corpus for reasoning, containing 92,000 verifiable examples across six domains: mathematics, code, science, logic, simulations, and tabular data. BBH~\cite{srivastava2023beyond} included 204 tasks spanning linguistics, child development, mathematics, commonsense reasoning, biology, physics, social bias, and software engineering. BBEH~\cite{kazemi2025big} replaced every BBH task with a new one testing similar reasoning skills but at significantly higher difficulty. OlympiadBench~\cite{he2024olympiadbench} is a bilingual, multimodal science benchmark containing 8,476 Olympiad-level math and physics problems, each accompanied by expert-authored step-by-step reasoning. Minerva Math~\cite{lewkowycz2022solving} consisted of 272 undergraduate-level STEM problems designed to test multi-step scientific reasoning in language models. SWE-bench~\cite{lewkowycz2022solving} included 2,294 software engineering tasks derived from real GitHub issues and corresponding pull requests across 12 popular Python repositories. SWE-bench Verified~\cite{yang2025swe} contained 50,000 instances collected from 128 GitHub repositories. SimpleQA~\cite{wei2024measuring} evaluated model performance on answering concise factual questions. LiveBench~\cite{White2025LiveBench} addressed concerns of test set contamination and human/model evaluation biases, covering diverse and challenging tasks across math, programming, reasoning, language, instruction-following, and data analysis. OJBench~\cite{wang2025ojbench} contained 232 competitive programming problems drawn from national and international contests (\textit{e.g.}, NOI and ICPC), providing a more rigorous test of reasoning under competitive conditions. AutoLogi~\cite{zhu2025autologi} and ZebraLogic~\cite{lin2025zebralogic} are dedicated to evaluating logical reasoning. $\tau^2$-Bench~\cite{barres2025tau} and ACEBench~\cite{chen2025acebench} are two complementary benchmarks designed to assess multi-turn tool usage. MultiChallenge~\cite{sirdeshmukh2025multichallenge} evaluated the ability of LLMs to engage in multi-turn dialogue with human users. MMReason~\cite{yao2025mmreason} encompassed complex problems spanning multiple domains and difficulty levels—ranging from pre-university to higher education, and from foundational to competition-level tasks—all of which require multi-step reasoning to solve. PolyMath~\cite{wang2025polymath} is a multilingual benchmark for mathematical reasoning across 18 languages and four difficulty levels. AMC 2023, AIME 2024/2025, CNMO 2024, and HMMT 2025 are high-difficulty mathematical competition benchmarks commonly used in the era of advanced reasoning models. MATH500 is a 500-question subset sampled from the full MATH~\cite{hendrycks2021measuring} benchmark. Humanity's Last Exam (HLE)~\cite{phan2025humanity} is a multimodal benchmark situated at the frontier of human knowledge, aiming to be the final comprehensive academic evaluation of its kind. HLE contains 2,500 questions spanning dozens of disciplines, including mathematics, the humanities, and natural sciences. Developed in collaboration with global experts, it includes both multiple-choice and short-answer formats suitable for automatic evaluation. Each question has a clear and verifiable answer that is not easily searchable online. Cutting-edge language models currently exhibit relatively low accuracy and calibration on this benchmark. PhyX~\cite{shen2025phyx} introduced a large-scale multimodal benchmark covering six physical domains and reasoning types, with tasks built from real visual scenes requiring image understanding, physical modeling, and symbolic reasoning, revealing comprehension bias and weak visual grounding in current MLLMs. Table \ref{tab:Performance} contrasts several well-known reasoning LLMs across tasks and capacities.
\begin{table}[tbp]
\centering
\caption{The Performance of Some Well-Known Reasoning Large Language Models on Test Benchmarks, respectively. The role of these test benchmarks is to compare the performance of mainstream reasoning models in terms of general tasks, alignment, mathematics/programming, and logical reasoning benchmarks. The * denotes the 14-language version.}
\vspace{-1mm}
\label{tab:Performance}

\makeatletter
\@ifundefined{tabincell}{\newcommand{\tabincell}[2]{\shortstack[#1]{#2}}}{}
\makeatother

\makeatletter
\@ifundefined{toprule}{\newcommand{\toprule}{\hline}}{}
\@ifundefined{midrule}{\newcommand{\midrule}{\hline}}{}
\@ifundefined{bottomrule}{\newcommand{\bottomrule}{\hline}}{}
\makeatother

\begingroup
\small
\setlength{\tabcolsep}{3pt}
\renewcommand{\arraystretch}{1.1}

\resizebox{0.97\textwidth}{!}{
\begin{tabular}{@{}clccccc@{}}
\toprule
&  & \textbf{OpenAI-o1~\cite{jaech2024openai}} & \textbf{DeepSeek-R1~\cite{guo2025deepseek}} & \tabincell{c}{\textbf{Grok-3-Beta}\\\textbf{(Think)}~\cite{xai2025grok3}} & \textbf{Gemini2.5-Pro~\cite{comanici2025gemini}} & \textbf{Qwen3-235B-A22B~\cite{yang2025qwen3}} \\
\midrule
& Architecture & -- & MoE & -- & -- & MoE \\
& \# Activated Params & -- & 37B & -- & -- & 22B \\
& \# Total Params & -- & 671B & -- & -- & 235B \\
\midrule
\multirow{4}{*}{\tabincell{c}{\textit{General}\\\textit{Tasks}}} & MMLU-Redux~\cite{gema2024we} & 92.8 & \underline{92.9} & -- & \textbf{93.7} & 92.7 \\
& MMMLU$^{*}$~\cite{hendrycks2020measuring} & \textbf{88.4} & 86.4 & -- & \underline{86.9} & 84.3 \\
& GPQA-Diamond~\cite{rein2024gpqa} & 78.0 & 71.5 & \underline{80.2} & \textbf{84.0} & 71.1 \\
& LiveBench~\cite{White2025LiveBench} & 75.7 & 71.6 & -- & \textbf{82.4} & \underline{77.1} \\
\midrule
\multirow{5}{*}{\tabincell{c}{\textit{Alignment}\\\textit{Tasks}}} & IFEval~\cite{zhou2023instruction} & \textbf{92.6} & 83.3 & -- & \underline{89.5} & 83.4 \\
& Arena-Hard~\cite{li2024crowdsourced} & 92.1 & 92.3 & -- & \textbf{96.4} & \underline{95.6} \\
& AlignBench v1.1 ~\cite{liu-etal-2024-alignbench}& 8.86 & 8.76 & -- & \textbf{9.03} & \underline{8.94} \\
& Creative Writing v3~\cite{creative-writing-bench-v3} & 81.7 & \underline{85.5} & -- & \textbf{86.0} & 84.6 \\
\midrule
\multirow{5}{*}{\tabincell{c}{\textit{Math\&Coding}\\\textit{Reasoning}}} & MATH-500~\cite{hendrycks2021measuring} & 96.4 & 97.3 &  & \textbf{98.8} & \underline{98.0} \\
& AIME'24 & 74.3 & 79.8 & 83.9 & \textbf{92.0} & \underline{85.7} \\
& AIME'25 & 79.2 & 70.0 & 77.3 & \textbf{86.7} & \underline{81.5} \\
& PolyMath~\cite{wang2025polymath} & 38.9 & 47.1 & -- & \underline{52.2} & \textbf{54.7} \\
& LiveCodeBench v5~\cite{jain2024livecodebench} & 63.9 & 64.3 & \underline{70.6} & 70.4 & \textbf{70.7} \\
\midrule
\multirow{2}{*}{\tabincell{c}{\textit{Logic}\\\textit{Reasoning}}} & ZebraLogic~\cite{lin2025zebralogic} & \underline{81.0} & 78.7 & -- & \textbf{87.4} & 80.3 \\
& AutoLogi~\cite{zhu2025autologi} & 79.8 & \underline{86.1} & -- & 85.4 & \textbf{89.0} \\
\bottomrule
\end{tabular}
}
\endgroup

\end{table}

\section{Open-source Tools and Frameworks}~\label{sec6}
VeRL~\cite{sheng2025hybridflow} is a system framework for efficient RLHF training and scheduling that integrates single- and multi-controller paradigms with a hierarchical interface, introduces a 3D-HybridEngine for parameter re-sharding, and applies automatic device mapping to optimize flexibility and resource utilization. TRLX~\cite{havrilla2023trlx} supported a wide range of distributed training paradigms, including data parallelism, model sharding, tensor parallelism, sequence parallelism, and pipeline parallelism. RL4LMs~\cite{ramamurthy2022reinforcement} is designed for optimizing language generators via reinforcement learning. The library implements online policy optimization algorithms and is compatible with any encoder or encoder–decoder model from the HuggingFace Transformers library~\cite{wolf2020transformers}, while supporting arbitrary reward functions. Colossal-AI was among the first to open-source a complete RLHF pipeline, i.e., ColossalChat~\cite{you2023colossalchat}, which included supervised data collection, supervised fine-tuning, reward model training, and reinforcement learning fine-tuning. DeepSpeed-Chat~\cite{yao2023deepspeed} integrates multiple optimization techniques for both training and inference into a unified framework. OpenRLHF~\cite{hu2024openrlhf} is built using Ray~\cite{liang2018rllib}, vLLM, DeepSpeed~\cite{yao2023deepspeed}, and HuggingFace Transformers~\cite{wolf2020transformers}, offering high resource efficiency and support for multiple training strategies. TRL~\cite{vonwerra2022trl} is designed for post-training foundation models using techniques such as SFT, PPO, and DPO. Built on top of the Transformers ecosystem~\cite{wolf2020transformers}, it supported various model architectures and modalities and scales across diverse hardware environments. Wang \textit{et al.}~\cite{wang2025ragen} proposed the RAGEN system, which enhances large language models' reasoning and decision-making capabilities in multi-turn interaction environments through the StarPO reinforcement learning framework. Fu \textit{et al.}~\cite{fu2025areal} introduced AReaL, a fully asynchronous reinforcement learning system that completely decouples the generation and training processes. In AReaL, rollout workers continuously generate new outputs without waiting, while training workers update the model immediately upon collecting a batch of data. ROLL~\cite{wang2025reinforcement} is a library designed to simplify reinforcement learning for large language models. It addresses the challenges faced by technologists, product developers, and algorithm researchers in managing multi-model, multi-stage training workflows. Nemo RL~\cite{nemo-rl} is a scalable and efficient post-training library capable of supporting models ranging from small to over 100 billion parameters and training environments from a single GPU to thousands. LlamaRL~\cite{wu2025llamarl} is a PyTorch-based distributed asynchronous reinforcement learning framework that enables efficient training of large-scale language models (ranging from 8B to 405B parameters), achieving significant speedups while maintaining strong performance. Yao \textit{et al.}~\cite{yao2025offpolicy} proposed Flash-LLM-RL, a package that patches vLLM to support model quantization with parameter updates. FlashRL~\cite{liu2025flashrl} introduced Truncated Importance Sampling (TIS) to mitigate the gap between rollout and training, enabling the use of quantized rollouts without sacrificing downstream performance. DistFlow~\cite{wang2025distflow} introduced a fully distributed reinforcement learning training framework that addresses the common single-controller bottleneck in large-scale language model post-training. It employed a multi-controller architecture and user-defined DAG-based task pipelines to achieve decentralized management of both data and computation.
\section{Open Discussion}~\label{sec7}
\subsection{Research Challenges}
While reinforcement learning has undeniably enhanced LLM alignment and reasoning, several fundamental challenges continue to hinder its full potential. 
\subsubsection{Scalability and Training Stability}
At the system level, large-scale RL on LLMs remains compute-intensive and sometimes unstable. Fine-tuning billion-parameter, high-action-space models demands vast resources and careful hyperparameter control; even with distributed frameworks like VeRL~\cite{sheng2025hybridflow}, achieving stable large-scale convergence is non-trivial. Misspecified rewards or poorly managed dynamics can cause policy collapse or divergence~\cite{yu2025dapo, rafailov2023direct}. Tooling is fragmented—libraries vary in interfaces and scope, complicating pipeline integration~\cite{havrilla2023trlx, yao2023deepspeed}. More efficient algorithms and robust, unified open-source frameworks are still needed to make RL both accessible and reliable at scale.
\subsubsection{Reward Design and Credit Assignment}

From a methodological perspective, challenges center on reward design, credit assignment, and exploration. Outcome-only rewards bias learning toward obvious, high-probability reasoning and overlook complex or unconventional solutions~\cite{wu2025invisible, yue2025does}. Richer signals, such as step-level dense feedback~\cite{zhang2025r1}, and entropy-based or diversity-promoting rewards~\cite{cui2025entropy, wang2025beyond}, show promise but remain immature; balancing exploration with efficient convergence is unresolved. Long-horizon credit assignment is especially difficult when rewards arrive only after lengthy reasoning~\cite{li2024larm}, motivating new reward schemes and algorithms.
\subsubsection{Theoretical Understanding and Reliability}

At a theoretical and analytical level, we lack a clear account of generalization and stability in RL-trained LLMs. It remains open whether RL genuinely yields new reasoning abilities or simply amplifies pre-trained patterns~\cite{zhao2025echo, yue2025does, wu2025invisible}. Misconfigured optimization can degrade calibration or core knowledge~\cite{korbak2022reinforcement, kotha2023understanding}. We need criteria for when RL helps versus hurts, and techniques, e.g., reward model regularization or conservative updates to curb instability. Deeper theory and lifecycle-wide interpretability studies are limited but essential for safer, more effective RL.
\subsubsection{Application-Level Challenges of Agentic LLMs}
At the application level, integrating LLMs with agentic and tool use via RL presents both exciting opportunities and unresolved difficulties. Recent work has started to treat LLMs as autonomous agents~\cite{peiyuan2024agile, feng2025group, wang2025spa, feng2025retool} that can plan, act, and interact with external tools or environments to accomplish complex goals. Reinforcement learning is a natural fit for training such agentic LLMs because it provides a feedback loop for trial-and-error learning in interactive scenarios. However, scaling this idea up surfaces challenges in efficiency, safety, and controllability. Training an LLM agent through environment interactions (e.g., simulated tool APIs~\cite{li2025torl}, web browsing\cite{dong2025tool}, or games~\cite{li2024larm}) is extremely resource-intensive, as it requires running the expensive model many times to explore different action sequences. Ensuring the safety of agentic behavior is even more critical: an RL-driven agent might discover strategies that technically maximize reward while violating user intent or ethical normswang~\cite{wang2025beyond, fu2025reward, miao2024inform, miao2025energy}. Unlike constrained single-turn text generation, an autonomous LLM agent could take a sequence of harmful or undesirable actions if its reward function is misspecified~\cite{wang2025ragen}. Therefore, developing safe RL techniques for LLM-based agents is an urgent area of research. Additionally, current approaches often lack a memory or planning mechanism to handle very long interaction sequences, making it hard for agents to perform tasks requiring long-term planning or to recover from mistakes~\cite{tan2025prospect, yu2025memagent, long2025seeing}. Integrating external memory combined with RL is a promising direction to address long-horizon agency, but it remains largely unexplored. Correspondingly, significant challenges remain with respect to datasets and evaluation benchmarks. Current studies often rely on bespoke datasets or task-specific benchmarks, making it difficult to systematically compare RL fine-tuning methods and to separate genuine improvements from task-specific gains. Although initial attempts such as Polymath~\cite{wang2025polymath}, Humanity's Last Exam~\cite{phan2025humanity} provide more rigorous tests for advanced reasoning, the broader field still lacks standardized, community-wide benchmarks and unified metrics. Developing such resources remains essential for establishing a solid theoretical and empirical foundation for RL-augmented LLMs.
\subsection{Future Trends}
\subsubsection{Evolving Learning Paradigms}
Looking ahead, we anticipate several key research trends to shape the future of RL-enhanced LLMs. First, there will be a push toward richer and more nuanced reward modeling. Rather than relying solely on outcome-based reward signals, future work is expected to incorporate process-level supervision and intermediate rewards that evaluate the quality of reasoning steps, justification logic, or adherence to constraints throughout the generation process~\cite{zhang2025r1, yang2025treerpo}. Such process-oriented rewards would help address the long-horizon credit assignment problem and encourage the model to develop more transparent and verifiable reasoning paths. Second, we foresee tighter integration of RL with structured reasoning paradigms and knowledge representations~\cite{luo2025graph}. By embedding logical or graph-structured inductive biases into the RL process, models may learn reasoning strategies that transfer more robustly to new tasks, as opposed to the relatively unstructured trial-and-error approach currently prevalent. For example, an LLM might use RL to learn how to traverse and update a knowledge graph or to plan sequences of tool calls, thereby gaining a form of systematic reasoning that purely neural approaches struggle with. 
\subsubsection{Expanding Application Frontiers}
Finally, the scope of RL applications for LLMs will continue to broaden, driving further innovations. We expect significant growth in multimodal reasoning tasks where LLMs augmented with vision, audio, or other modalities use RL to coordinate between modalities and achieve complex goals~\cite{huang2025vision, ghosh2025visual} as well as in specialized domains like scientific research assistants, formal theorem proving, or decision-support systems. Each of these new domains will bring its own challenges, likely necessitating customized reward functions and safety considerations. The introduction of more comprehensive benchmarks and competitions targeting these scenarios will spur methodological advances by highlighting the limitations of current techniques. Meanwhile, given that agentic systems are inherently well-suited to the RL training paradigm and hold broad application prospects, RL-enhanced Agentic LLMs are undoubtedly an emerging and future technological trend. 
\subsubsection{Toward a Virtuous Research Cycle}
In combination, these trends indicate a shift beyond today’s relatively conservative fine-tuning toward a paradigm of using RL to train more adaptive, robust, and safe LLMs. The long-term vision is that reinforcement learning, underpinned by strong theoretical insights and practical tools, will enable LLMs to not only align with human values but also continuously improve their reasoning through experience, ultimately inching closer to systems that can learn how to reason in a human-like, self-correcting manner. Each challenge outlined above also represents an opportunity: by overcoming issues of stability, reward design, theoretical understanding, and evaluation, the field can unlock the next wave of progress in large-scale intelligent systems. Each emerging solution, in turn, feeds into a virtuous cycle—better tools and benchmarks lead to more rigorous research, which yields more capable and aligned models, which then require new evaluation standards—pushing the frontier of what RL-enhanced LLMs can achieve.
When considered together, these trends signify a shift toward more comprehensive, structured, and diversified RL training for LLMs. Each challenge represents both a limitation and an opportunity: resolving scalability, reward design, theory gaps, application challenges of agentic LLMs, and evaluation will unlock new reasoning frontiers. Crucially, solutions reinforce each other better tools and benchmarks lead to stronger models, which in turn necessitate improved evaluation. This self-reinforcing cycle will drive the evolution of more aligned, generalizable, and safe RL-enhanced LLMs.
\section{Conclusion}~\label{sec8}
This survey presents a comprehensive review of reinforcement learning for large language models, organized around the full training lifecycle from pre-training to alignment and reasoning. Particular emphasis is given to RLVR technology, which represents a promising direction for incorporating objective and reliable optimization signals. In addition, the survey consolidates datasets, benchmarks, and open-source frameworks, providing a structured reference for both evaluation and practical implementation. By integrating these perspectives, the survey delivers a lifecycle-based synthesis that highlights both methodological advances and supporting resources, serving as a state-of-the-art reference for future research in RL-enhanced LLMs.

\bibliographystyle{ACM-Reference-Format}
\bibliography{survey.bib}

\appendix

\end{document}